\documentclass[11pt]{article}

\usepackage[margin=1in]{geometry}
\usepackage{setspace}
\usepackage{float}
\usepackage{caption}
\usepackage{parskip}

\usepackage{newtxtext,newtxmath}
\usepackage{microtype}

\usepackage{graphicx}
\usepackage{placeins}
\usepackage{amsmath}
\usepackage{multicol}
\renewenvironment{abstract}
{
  \section*{Abstract}
}
{
  \par
}

\usepackage[
  backend=biber,
  style=numeric-comp,
  sorting=none,
  maxnames=1,
  autocite=superscript
]{biblatex}
\let\cite\autocite

\usepackage[
  colorlinks=true,
  citecolor=blue,
  linkcolor=blue,
  urlcolor=blue
]{hyperref}

\title{\textbf{VitaminP: cross-modal learning enables whole-cell segmentation from routine histology}}

\author{
Yasin Shokrollahi\textsuperscript{1,2},
Karina B. Pinao Gonzales\textsuperscript{1,2},
Elizve N. Barrientos Toro\textsuperscript{1,2},
Paul Acosta\textsuperscript{1,2},\\
Patient Mosaic Team\textsuperscript{\#3},
Pingjun Chen\textsuperscript{1,2*},
Yinyin Yuan\textsuperscript{1,2*},
Xiaoxi Pan\textsuperscript{1,2*}
}

\date{}

\begin{document}

\maketitle

\vspace{0.8em}
\noindent\textsuperscript{1}Department of Translational Molecular Pathology, Division of Pathology and Laboratory Medicine,\\
The University of Texas MD Anderson Cancer Center, Houston, TX, USA.\\
\textsuperscript{2}Institute for Data Science in Oncology,\\
The University of Texas MD Anderson Cancer Center, Houston, TX, USA.\\
\textsuperscript{3}The University of Texas MD Anderson Cancer Center, Houston, TX, USA.\\[0.4em]
\textsuperscript{\#}Full collaborator list provided in the Supplementary Information.\\
\textsuperscript{*}Corresponding authors.

\vspace{1em}

% ======================
% Abstract
% ======================
\begin{abstract}
\noindent
Accurate whole-cell and nuclear segmentation is essential for precision pathology and spatial omics, yet routine hematoxylin and eosin (H\&E) staining provides limited cytoplasmic contrast, restricting analyses to nuclei. Multiplex immunofluorescence (mIF) facilitates precise whole-cell delineation but remains constrained by cost and accessibility. We introduce VitaminP, a cross-modal learning framework enabling whole-cell segmentation from H\&E images. By learning from paired H\&E--mIF data, VitaminP transfers molecular boundary information from mIF to overcome cytoplasmic contrast in H\&E, establishing cross-modal supervision as a general strategy for recovering missing biological structure. We train VitaminP on 14 public datasets covering 34 cancer types and over 7 million instances, integrating publicly available labels with extensive annotations generated in this study, forming one of the largest resources for segmentation. VitaminP outperforms four state-of-the-art methods and generalizes to unseen datasets, including an in-house dataset spanning 24 rare cancer types. We further developed VitaminPScope, an open-source platform providing an interface for scalable inference and enabling broad adoption\footnote{
The source code and platform are publicly available at:
\url{https://github.com/idso-fa1-pathology/VitaminP} and
\url{https://github.com/idso-fa1-pathology/VitaminPScope}.
}.
\end{abstract}

% ======================
% Introduction
% ======================
\section{Introduction}

Whole-cell and nuclear segmentation is fundamental to computational pathology, enabling cell phenotyping, spatial modeling of the tumor microenvironment, and biomarker quantification\autocite{mahbod_nuinsseg_2024,debsarkar_advancements_2025,mcgenity_artificial_2024}. In routine clinical workflows, hematoxylin and eosin (H\&E) staining remains widely accessible, cost-effective, and rich in morphological detail. However, its limited cytoplasmic contrast fundamentally prevents precise delineation of cell membranes, hindering accurate whole-cell segmentation and quantification. As a result, most downstream analyses rely on nuclear features rather than full cellular morphology. Multiplex immunofluorescence (mIF) addresses this limitation by enabling simultaneous detection of multiple protein biomarkers in the cytoplasm and on the membrane within a single tissue section, facilitating precise delineation of cellular boundaries, accurate identification of cell types, and detailed analysis of the spatial organization\cite{rojas_multiplex_2022,yosofvand_spatial_2024}.

Recent advances in vision transformers (ViT) and self-supervised foundation models have improved pathology image analysis by learning representations from large-scale datasets\cite{komura_machine_2025,muller-franzes_medical_2025,oquab_dinov2_2023,xu_vision_2024,vorontsov_foundation_2024,tang_prototype-based_2025,abimouloud_advancing_2025,ignatov_histopathological_2024,burlingame_shift_2020}, but remain limited in recovering missing cellular boundaries from routine stains. The growing availability of paired H\&E--mIF datasets has motivated the development of cross-modal learning strategies\cite{schroder_hemit_2025,lin_highly_2018}. Virtual staining approaches, including SHIFT\cite{burlingame_shift_2020}, DeepLIIF\cite{ghahremani_deep_2022}, HEMIT\cite{schroder_hemit_2025}, ROSIE\cite{wu_rosie_2025}, and MIPHEI-ViT\cite{balezo_miphei-vit_2025}, use generative models to infer molecular marker expression from H\&E morphology. While these methods demonstrate correlations between tissue structure and protein expression, they primarily focused on image-to-image translation rather than cellular boundary delineation.

Whole-cell segmentation remains a persistent challenge across staining modalities. Even state-of-the-art generalist segmentation frameworks encounter distinct inherent constraints. Prompt-based adaptations such as CellSAM\cite{marks_cellsam_2025} required iterative inference, causing computational cost to grow linearly with cell count and posing challenges for whole-slide images (WSIs). In contrast, dense prediction models such as Cellpose-SAM\cite{pachitariu_cellpose-sam_2025} improved scalability by enabling direct pixel-wise segmentation in a single forward pass, avoiding iterative prompting; however, they do not explicitly exploit spatially aligned cross-modal supervision to transfer molecular boundary information across staining modalities. Similarly, frameworks built on SAM encoders\cite{kirillov_segment_anything_2023}, including CellViT++\cite{horst_cellvit_2025}, extract powerful visual features yet remain confined to a single modality. Consequently, existing methods rarely exploit the complementary molecular signals in mIF to resolve ambiguous cellular boundaries in H\&E, even though paired data offer a unique supervisory advantage. This gap likely reflects the scarcity of high-quality whole-cell annotations, owing to the substantial cost, time requirements, and inter-observer variability associated with manual labeling\cite{mahbod_nuinsseg_2024,moen_deep_2019,greenwald_whole-cell_2022}.

Here we present VitaminP (Vision transformer-assisted multimodal integration for Pathology) (Fig.~\ref{fig:fig1}a), a cross-modal learning framework for accurate whole-cell and nuclear segmentation by leveraging complementary information from paired H\&E--mIF data during training. To accommodate diverse needs, VitaminP includes three model variants: VitaminP-Dual employs a dual-encoder ViT architecture to jointly learn from aligned H\&E--mIF pairs, delivering simultaneous whole-cell and nuclear segmentation across modalities; VitaminP-Syn extends this capability to unpaired settings by generating synthetic mIF representations from H\&E images and integrating them within the joint pipeline; and VitaminP-Flex provides an efficient, modality-agnostic solution optimized for high-throughput deployment. High-quality whole-cell and nuclear annotations were rapidly curated using an AI--human collaboration pipeline on paired H\&E--mIF datasets (Fig.~\ref{fig:fig1}b). VitaminP was trained on 14 public datasets spanning 34 cancer types and over 7 million annotated instances (Fig.~\ref{fig:fig1}c--d; Extended Data Fig.~\ref{fig:ext_fig1}d), representing one of the largest publicly available resources for whole-cell and nuclear segmentation to date. Across multiple external validation datasets, VitaminP outperformed state-of-the-art methods overall. Its robustness was further validated against pathologist-curated annotations from 24 rare cancer types exhibiting high morphological heterogeneity. Importantly, cross-modal supervision substantially improves whole-cell boundary delineation in H\&E images, where cytoplasmic contours are often poorly resolved\cite{graham_hover-net_2019}. This enhanced H\&E-driven whole-cell segmentation enables more accurate spatial transcriptomic analysis and supports quantitative morphometric profiling, including nuclear-to-cytoplasmic (N/C) ratio measurements. To facilitate widespread adoption, we introduce VitaminPScope, an open-source platform that provides accessible, high-accuracy cell-level pathology analysis.

\section*{Results}

\subsection*{Cross-modal learning enabled by AI--human annotation curation}

To enable whole-cell segmentation from H\&E using molecular supervision from mIF, we designed VitaminP as a dense segmentation framework that predicts all instances in a single pass (Fig.~\ref{fig:fig1}a). Unlike prompt-based adaptations~\cite{marks_cellsam_2025,pachitariu_cellpose-sam_2025} or unimodal pipelines~\cite{horst_cellvit_2025,graham_hover-net_2019}, VitaminP learns from paired H\&E--mIF data using a transformer-based architecture with shared and modality-specific features. The framework comprises three variants for distinct application scenarios. VitaminP-Dual employs dual ViT-B/14 encoders~\cite{dosovitskiy2021} to jointly process spatially aligned H\&E--mIF images, enabling direct cross-modal feature fusion for boundary refinement. VitaminP-Syn extends this strategy to H\&E-only settings by synthesizing virtual mIF representations from H\&E inputs using a Pix2Pix-based~\cite{isola2017} translation module prior to cell segmentation. Complementing these two, VitaminP-Flex provides an efficient, modality-agnostic alternative using a single ViT-L/14 encoder~\cite{dosovitskiy2021} to support scalable inference across heterogeneous inputs. VitaminP-Dual and VitaminP-Syn were trained using spatially paired datasets to explicitly leverage multimodal information, whereas VitaminP-Flex was trained across all available datasets to enable modality-agnostic generalization without requiring paired inputs. Across all model variants, H\&E whole-cell segmentation was supervised using mIF-derived annotations.

Effective cross-modal learning requires large-scale supervision across diverse tissue morphologies. To efficiently curate whole-cell and nuclear annotations, we implemented an AI--human collaboration pipeline (Fig.~\ref{fig:fig1}b). Initial segmentation masks at WSI level were generated using modality-specific models, CellViT++~\cite{horst_cellvit_2025} for H\&E images to deliver nuclear annotations, while Cellpose-SAM~\cite{pachitariu_cellpose-sam_2025} was used for mIF images to derive both whole-cell and nuclear annotations. For paired data, mIF-derived whole-cell masks were transferred to the corresponding H\&E images and used for supervision during training. Pathologists then reviewed these predictions to identify high-confidence regions of interest (ROIs) with reliable whole-cell and nuclear segmentations, reducing the need for labor-intensive manual contour delineation. We applied the pipeline to 50 pairs of H\&E and mIF WSIs, including images from the ORION platform~\cite{lin_high-plex_2023} (n = 41 pairs) and the 10x Xenium platform with multimodal staining~\cite{10xgenomics_datasets} (n = 9 pairs). This pipeline produced 5,246 paired H\&E--mIF tiles (512 $\times$ 512 pixels), including 3,337 pairs from ORION (2,420,614 instance-level annotations) and 1,909 pairs from 10x Xenium (742,848 instance-level annotations) (Extended Data Fig.~\ref{fig:ext_fig1}a--c). ROIs were selected by two independent pathologists, with an average annotation time of 60 minutes per slide. To complement the paired datasets providing direct cross-modal supervision, we incorporated 12 additional public datasets, including TissueNet~\cite{greenwald_whole-cell_2022}, PanNuke~\cite{gamper_pannuke_2020}, Lizard~\cite{graham_lizard_2021}, MoNuSeg~\cite{kumar_multi-organ_2020}, TNBC~\cite{naylor_segmentation_2019}, NuInsSeg~\cite{mahbod_nuinsseg_2024}, CryoNuSeg~\cite{mahbod_cryonuseg_2021}, BC (Breast Carcinoma)~\cite{amgad_structured_2019}, CoNSeP~\cite{graham_hover-net_2019}, MoNuSAC~\cite{verma_monusac2020_2021}, Kumar~\cite{kumar_dataset_2017}, and CPM-17~\cite{vu_methods_2019}. Overall, the training data spanned 34 cancer types and multiple anatomical sites (Fig.~\ref{fig:fig1}c--d; Extended Data Fig.~\ref{fig:ext_fig1}d). This harmonized multi-source dataset represents one of the largest publicly curated resources for whole-cell and nuclear segmentation to date.

\subsection*{Paired supervision improves whole-cell segmentation}

To assess whether paired H\&E--mIF images provide complementary information, we visualized modality-specific embeddings prior to task-specific fine-tuning. UMAP exhibited a clear separation between H\&E and mIF embeddings, with each modality forming distinct clusters (Fig.~\ref{fig:fig1}e), supporting the rationale for cross-modal learning in downstream prediction tasks.

To evaluate the value of explicit paired H\&E--mIF supervision, we compared VitaminP-Dual and VitaminP-Syn with VitaminP-Flex on ORION and 10x Xenium datasets (Fig.~\ref{fig:fig2}a). Performance was assessed using the panoptic quality~\cite{kirillov_panoptic_2018} (PQ) metric across four tasks: H\&E-derived nuclear and whole-cell segmentation, and mIF-derived nuclear and whole-cell segmentation. On H\&E-derived tasks, the three models achieved comparable performance for nuclear segmentation (VitaminP-Dual: 0.68, VitaminP-Syn: 0.68, VitaminP-Flex: 0.70), while VitaminP-Dual and VitaminP-Syn substantially improved whole-cell segmentation (0.72 and 0.72) relative to VitaminP-Flex (0.52; Fig.~2a). This improvement reflects the limitation of brightfield staining, where nuclear boundaries are often visible but cytoplasmic contours remain ambiguous. This pattern was also observed in mIF-derived tasks, where all three models achieved comparable performance for both nuclear (VitaminP-Dual: 0.72, VitaminP-Syn: 0.71, VitaminP-Flex: 0.75) and cell (VitaminP-Dual: 0.71, VitaminP-Syn: 0.72, VitaminP-Flex: 0.73) segmentation (Fig.~\ref{fig:fig2}a), with VitaminP-Flex performing slightly better. This may reflect that the input modality and annotation source for VitaminP-Flex are mIF, whereas VitaminP-Dual and VitaminP-Syn also incorporate H\&E input, which may slightly reduce performance on mIF segmentation. Notably, the H\&E whole-cell reference masks were derived from paired mIF images. This evaluation scheme could favor the performance of VitaminP-Dual and VitaminP-Syn, which leverage multimodal information, whereas VitaminP-Flex was only fed with H\&E images for inference.

\vspace{1em}

Together, these findings indicate that cross-modal learning enhances segmentation performance, particularly for H\&E whole-cell delineation. Nevertheless, VitaminP-Flex, trained with mIF supervision, represents the more practical model for broad adoption, as paired H\&E--mIF data remain limited and virtual mIF synthesis approaches introduce an additional layer of model dependence that requires careful validation.

\begin{center}
    \includegraphics[width=\textwidth]{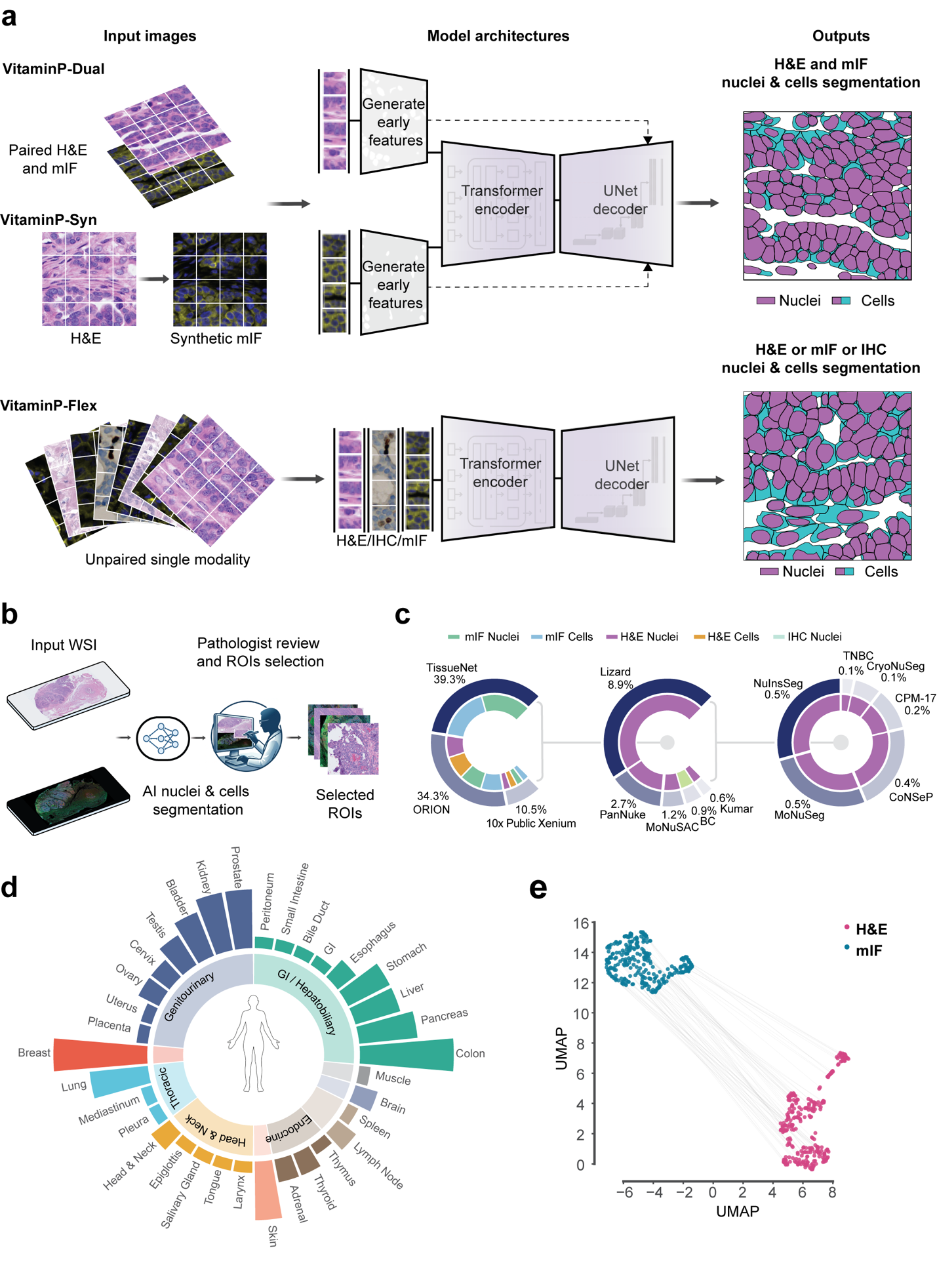}
\end{center}

\phantomsection
\captionof{figure}{
\textbf{Cross-modal whole-cell segmentation framework and multi-source dataset for VitaminP.}
\textbf{a,} Schematic overview of VitaminP and its three variants. VitaminP-Dual jointly processes spatially aligned H\&E--mIF images. VitaminP-Syn extends this framework to H\&E-only inputs by generating synthetic mIF representations. VitaminP-Flex employs a single-encoder design optimized for modality-agnostic inference across heterogeneous inputs (H\&E, mIF, or IHC).
\textbf{b,} AI--human collaboration annotation pipeline for paired datasets. WSIs are initially segmented using modality-specific models, followed by pathologist review and selection of high-confidence ROIs for whole-cell and nuclear annotation.
\textbf{c,} Composition of the harmonized multi-source dataset. Fourteen public datasets spanning 34 cancer types contribute over 7 million annotated nuclei and cells.
\textbf{d,} Organ and cancer-type distribution of the training dataset.
\textbf{e,} UMAP visualization of pretrained transformer embeddings prior to task-specific fine-tuning. Although H\&E and mIF images occupy distinct spectral domains, semantically corresponding cellular structures cluster closely in feature space, supporting the suitability of the pretrained backbone for cross-modal learning.
}
\label{fig:fig1}

\subsection*{Benchmarking VitaminP-Flex's performance on public datasets}

To evaluate segmentation performance, we benchmarked VitaminP-Flex against four widely used models, CellSAM~\cite{marks_cellsam_2025}, Cellpose-SAM~\cite{pachitariu_cellpose-sam_2025}, Hover-Net~\cite{graham_hover-net_2019}, and CellViT++~\cite{horst_cellvit_2025}, across 12 public datasets spanning diverse tissues and modalities (Fig.~\ref{fig:fig2}b--c). All models were evaluated on held-out test sets under the same conditions (see Methods). 
\vspace{1em}

Across H\&E datasets, VitaminP-Flex achieved the highest average PQ (0.62), outperforming Hover-Net (0.51), CellViT++ (0.52), and CellSAM (0.49), and comparable to Cellpose-SAM (0.61) (Fig.~\ref{fig:fig2}b). VitaminP-Flex further generalized to multiplex immunohistochemistry (IHC) and IF images, achieving competitive performance across modalities (Fig.~\ref{fig:fig2}b). Collectively, VitaminP-Flex most frequently achieved Top-1 performance (8/13) and consistently ranked within the Top-2 or Top-3, indicating robust performance without dataset-specific tuning, consistent with previous observations~\cite{marks_cellsam_2025} (Fig.~\ref{fig:fig2}c--d; Extended Data Fig.~\ref{fig:ext_fig2}).
\vspace{1em}

To evaluate generalization, we tested VitaminP-Flex on three unseen datasets under a strict zero-shot setting (Fig.~\ref{fig:fig2}e). For H\&E nuclei segmentation, VitaminP-Flex achieved strong performance on CPM-15 and PanopTILs~\cite{liu_panoptic_2024}, outperforming CellSAM on CPM-15 and remaining comparable to Cellpose-SAM across both datasets. On the mixed-modality benchmark, the DSB2018 dataset, VitaminP-Flex achieved lower performance than Cellpose-SAM and CellSAM; however, both models were pretrained on datasets overlapping with DSB2018, whereas VitaminP-Flex was evaluated without exposure to it during training. Across all datasets, VitaminP-Flex consistently outperformed Hover-Net and CellViT++, which were restricted to H\&E images by design (Fig.~\ref{fig:fig2}e).
\vspace{1em}

To evaluate computational efficiency, we benchmarked VitaminP-Flex against CellSAM and Cellpose-SAM across five metrics: single-tile latency, whole-slide processing time, peak GPU memory usage, segmentation throughput, and area processing rate (Fig.~\ref{fig:fig2}f). Using CellSAM as the 1$\times$ baseline, VitaminP-Flex achieved substantial improvements in speed and throughput, including faster tile processing (1.93$\times$) and higher whole-slide (17.61$\times$) and segmentation throughput (20.41$\times$). Although VitaminP-Flex exhibited lower memory efficiency than Cellpose-SAM (1.12$\times$ vs 2.18$\times$ relative to the CellSAM baseline), this reflects its dense single-pass design, which processes all cells simultaneously rather than iteratively. Overall, these results show that VitaminP-Flex provides a scalable approach that maintains strong segmentation performance across diverse benchmarks, reducing the need for dataset-specific model selection.

\subsection*{VitaminP-derived whole-cell segmentation from H\&E is comparable to marker-guided approaches}

Whole-cell annotations for H\&E images are largely absent from existing benchmarking datasets. To enable a rigorous evaluation of whole-cell segmentation performance, we conducted two independent validation studies with pathologist annotations (Fig.~\ref{fig:fig3}).
\vspace{1em}

In the first assessment, we evaluated performance on the paired ORION and 10x Xenium datasets. Twenty image tiles (512 $\times$ 512 pixels) were randomly selected from the test split and reviewed by pathologists, who annotated whole-cell instances with guidance from paired mIF images when necessary (Fig.~\ref{fig:fig3}a--b). However, consistent with the intrinsic limitations of cytoplasmic boundary delineation, annotations were restricted to cells whose contours could be identified with high confidence. This conservative strategy ensures that the reference masks include only reliably delineated cells, minimizing uncertainty introduced by ambiguous boundaries. On this curated set (Fig.~\ref{fig:fig3}a), VitaminP variants generated whole-cell predictions directly from H\&E images, while Cellpose-SAM~\cite{pachitariu_cellpose-sam_2025} and CellSAM~\cite{marks_cellsam_2025} produced whole-cell segmentations from mIF inputs, as they were not designed to infer whole-cell boundaries from H\&E. Despite this input advantage for Cellpose-SAM and CellSAM, VitaminP-Dual achieved the highest performance (PQ: 0.42), showing strong agreement with pathologist-defined cell boundaries. Because cytoplasmic boundaries were annotated only where cell contours could be identified with high confidence (Fig.~\ref{fig:fig3}b), the reference masks represent a subset of reliably delineated cells. Consequently, correctly predicted but unlabeled cells are counted as false positives, which may reduce evaluation metrics.

\vspace{1em}

In the second validation study, we assessed robustness under pronounced morphological shift by curating an independent rare cancer dataset (RC-24), comprising 24 image tiles (512 $\times$ 512 pixels), each representing a distinct tumor subtype absent from the training data (Fig.~\ref{fig:fig3}c). Nuclear annotations were exhaustively performed across all tiles with DAPI channel guidance to ensure accurate boundary delineation and identification, particularly in regions with overlapping structures that may appear merged in H\&E. Whole-cell annotations were generated with mIF guidance when needed. However, cytoplasmic boundary delineation depends on marker specificity and signal intensity; therefore, pathologists annotated only cells whose contours could be identified with high confidence. Accordingly, the RC-24 dataset comprises 11,798 manually delineated objects, including 7,646 nuclei and 4,152 high-confidence whole-cell instances across 24 rare tumor types (Fig.~\ref{fig:fig3}c). For whole-cell segmentation, VitaminP-Dual achieved a mean PQ of 0.40, VitaminP-Syn achieved a mean PQ of 0.35, and VitaminP-Flex achieved a mean PQ of 0.41 (Fig.~\ref{fig:fig3}d). In comparison, the mIF-based methods achieved mean PQ scores of 0.40 (CellSAM) and 0.43 (Cellpose-SAM) (Fig.~\ref{fig:fig3}d). Similarly, our conservative annotation strategy may contribute to lower evaluation metrics (Fig.~\ref{fig:fig3}e).

\begin{center}
    \includegraphics[width=\textwidth]{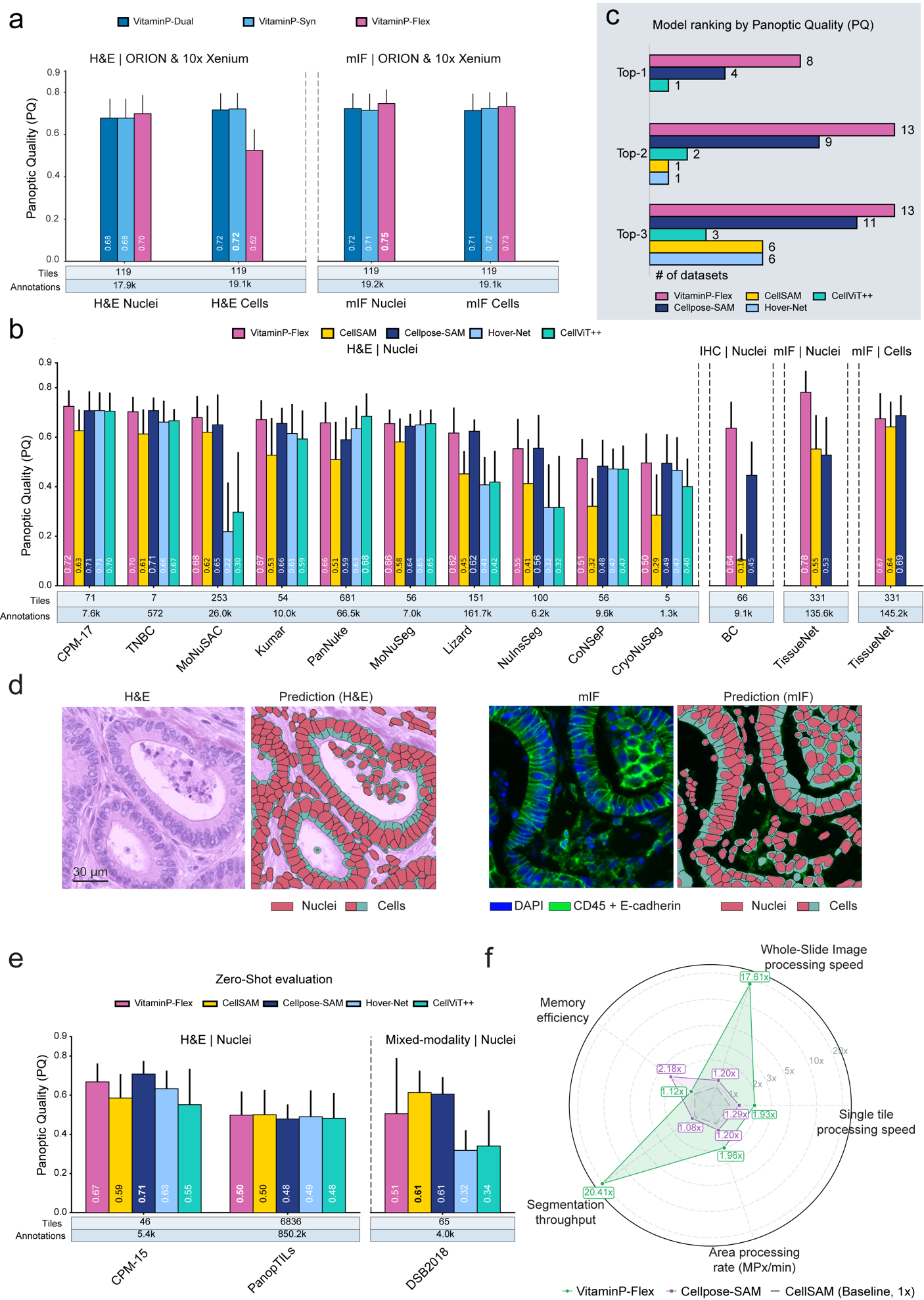}
\end{center}
\captionof{figure}{\textbf{Performance benchmark of VitaminP and comparison methods on nuclear and whole-cell segmentation.} \textbf{a,} Paired-modality evaluation on ORION and 10x Xenium datasets. Bars represent mean PQ; error bars denote s.d. \textbf{b,} Benchmarking of VitaminP-Flex across 12 public datasets spanning H\&E, IHC, and mIF modalities. VitaminP-Flex shows strong performance relative to CellSAM, Cellpose-SAM, Hover-Net, and CellViT++ across diverse tissue types and acquisition protocols. Tile and annotation counts are reported below each dataset. \textbf{c,} Model ranking frequency across public benchmarks. Number of datasets in which each model ranks within Top-1, Top-2, or Top-3 by PQ, indicating the consistency and generalization of VitaminP-Flex relative to competing methods. \textbf{d,} Representative qualitative results for H\&E and mIF nuclear and whole-cell segmentation. Scale bar, 30 $\mu$m. \textbf{e,} Zero-shot evaluation on external datasets (CPM-15, PanopTILs, DSB2018). \textbf{f,} Whole-slide scalability comparison (log scale). Metrics are normalized to CellSAM single-task inference (1$\times$ baseline), include single-tile speed, whole-slide processing speed, peak GPU memory usage, segmentation throughput, and area processing rate.}
\label{fig:fig2}

\vspace{1em}

To further contextualize boundary performance without exhaustive expert annotations, we quantified cross-modal similarity by aligning H\&E-derived cell segmentations with mIF-derived segmentations on the same tiles (Fig.~\ref{fig:fig3}f). VitaminP-Dual exhibited high concordance with mIF pipelines (Dice 0.90 vs.\ CellSAM (mIF); 0.87 vs.\ Cellpose-SAM (mIF)). VitaminP-Flex (H\&E) showed similarly strong agreement (Dice 0.82 vs.\ CellSAM (mIF); 0.81 vs.\ Cellpose-SAM (mIF)), indicating that H\&E-derived whole-cell segmentation achieves performance comparable to marker-guided approaches.
\vspace{1em}

Together, these comparisons indicate that VitaminP-Flex recovers cell boundaries from H\&E alone that closely approximate those inferred from molecularly resolved mIF segmentation, reinforcing its robustness in settings where cytoplasmic delineation remains challenging.

\subsection*{VitaminP-Flex improves whole-cell segmentation for Xenium and enhances morphometric profiling}

To demonstrate the downstream utility of whole-cell segmentation derived from H\&E images, we applied VitaminP-Flex to H\&E images paired with single-cell spatial transcriptomics data generated by the 10x Xenium platform, specifically the version providing only a DAPI morphology image. The analysis included three cancer samples from breast, lung, and colorectal tissues. We benchmarked this approach against the default segmentation in the Xenium workflow, where cell masks are generated by isotropic nuclear expansion guided by DAPI morphology (Fig.~\ref{fig:fig4}a). As no ground-truth cell segmentation was available, we evaluated performance using downstream clustering based on reassigned transcripts (Fig.~\ref{fig:fig4}b; Extended Data Fig.~\ref{fig:ext_fig4}). Clustering performance was quantified using the Silhouette score across $K = 2$--$10$ clusters (Fig.~\ref{fig:fig4}c; Extended Data Fig.~\ref{fig:ext_fig4}). Across all three samples (breast, lung, and colorectal), VitaminP-Flex improved clustering performance relative to nuclei-expanded segmentation, although the magnitude of improvement varied by sample. At higher clustering resolution ($K = 10$), VitaminP-Flex consistently achieved higher Silhouette scores across samples (breast: 0.51 vs.\ 0.32; lung: 0.46 vs.\ 0.36; colorectal: 0.32 vs.\ 0.26), indicating more stable and well-separated clusters following transcript reassignment (Fig.~\ref{fig:fig4}b).

\begin{center}
    \includegraphics[width=\textwidth]{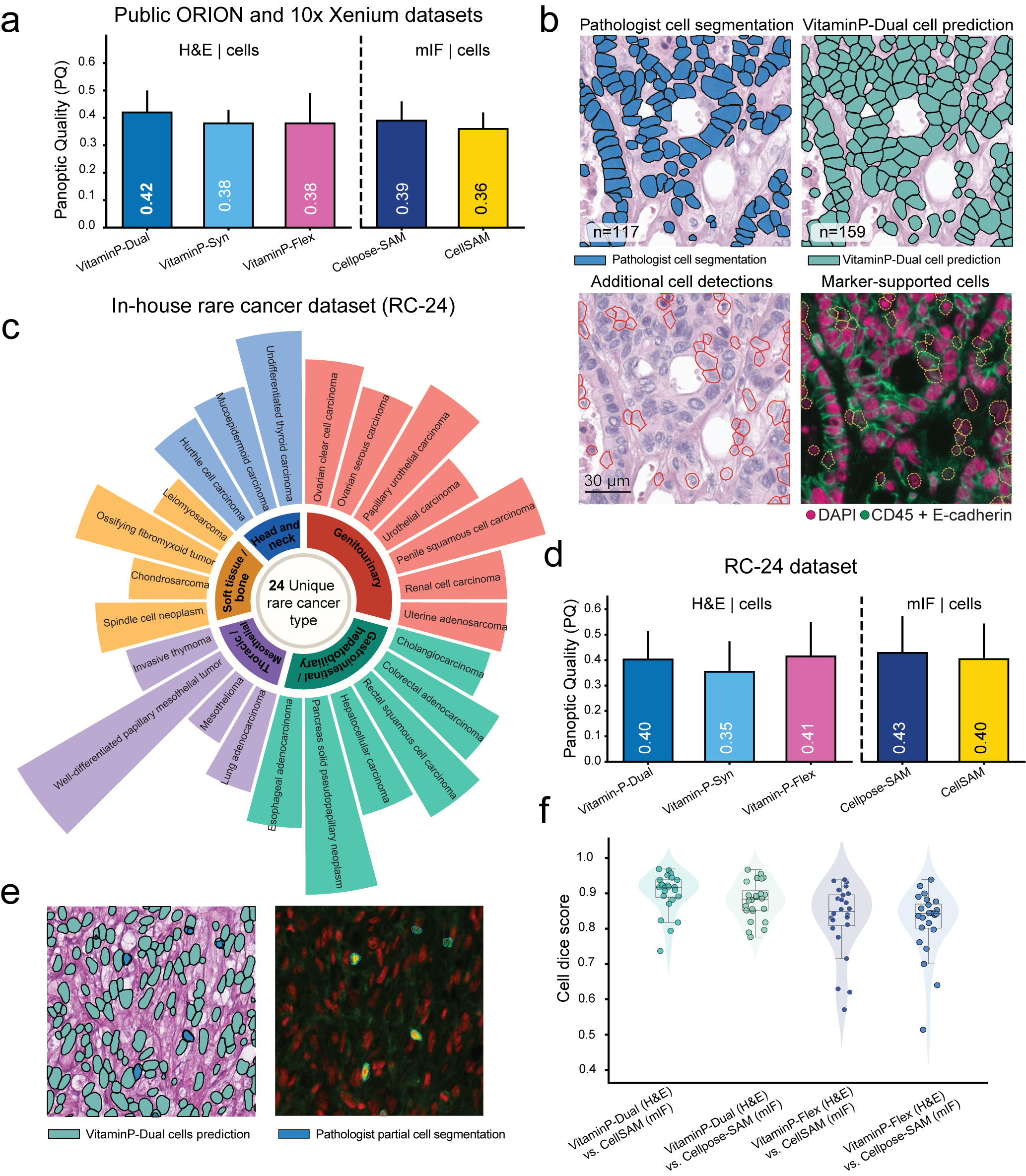}
\end{center}

\captionof{figure}{\textbf{Performance benchmark of VitaminP and comparison methods on whole-cell segmentation.} \textbf{a,} Whole-cell segmentation performance on held-out paired ORION and 10x Xenium tiles (n = 20; 512 $\times$ 512 px). Annotations were generated by expert pathologists with mIF guidance and restricted to high-confidence cells. \textbf{b,} Representative example comparing pathologist annotations, VitaminP-Dual predictions, and additional detected cells. Additional detections in H\&E correspond to cells supported by signal in the paired mIF image. \textbf{c,} Sunburst chart summarizing the distribution of RC-24 dataset, including 11,798 manually delineated instances (7,646 nuclei and 4,152 high-confidence whole-cell instances). \textbf{d,} Whole-cell segmentation performance evaluated against high-confidence partial annotations (H\&E). Panoptic Quality (PQ) is shown as the primary metric. \textbf{e,} Representative example of whole-cell segmentation under partial annotation. VitaminP-Dual predictions (red) are compared with high-confidence pathologist annotations (yellow), which cover only a subset of clearly delineated cells. The paired mIF/DAPI image (right) provides complementary information. \textbf{f,} Cross-modal consistency analysis of whole-cell segmentation on paired tiles. Dice agreement between H\&E-derived predictions and mIF-derived segmentations demonstrate strong geometric concordance, particularly for VitaminP-Dual. Points represent per-tile values.}
\label{fig:fig3}

\vspace{1em}

Beyond spatial transcriptomics, H\&E-derived whole-cell segmentation enables estimation of the nuclear-to-cytoplasmic (N/C) ratio and reveals spatial heterogeneity in cell morphology across tumor and stromal regions (Fig.~\ref{fig:fig4}d). To demonstrate scalability and biological relevance, we extended this analysis to a whole-slide image and combined segmentation outputs with cell-type classification (Extended Data Fig.~\ref{fig:ext_fig5}). Using CellViT++-based classification~\cite{horst_cellvit_2025}, we generated spatial maps of major cell types, including epithelial cells, stromal cells, tumor-infiltrating lymphocytes (TILs), and other cell populations. Integrating these labels with N/C ratio measurements enabled cell-type-specific morphometric analysis. Consistent with established histomorphology~\cite{komura_restaining-based_2023}, TILs exhibited higher N/C ratios relative to stromal and epithelial cells (two-sided Mann--Whitney U test, both $P < 0.0001$), whereas epithelial cells displayed broader variability.
\vspace{1em}

Together, these results demonstrate that whole-cell segmentation derived from H\&E images improves transcript assignment, enhances clustering structure, and enables large-scale morphometric analysis without requiring membrane markers at inference.

\subsection*{VitaminPScope enables interactive deployment for whole-slide images}

To facilitate deployment of VitaminP models for accessible pathology analysis, we developed VitaminPScope, an open-source web platform for whole-slide and ROI inference, visualization, and analysis of AI-derived segmentations (Fig.~\ref{fig:fig5}a). The platform integrates VitaminP models into a modular microservice architecture, providing scalable inference on gigapixel pathology images through an intuitive browser-based interface. VitaminPScope adopts a distributed architecture comprising three components: a web-based user interface, an application backend, and a dedicated AI inference service. Whole-slide images (WSIs) can be uploaded through the browser interface, with support for common formats including SVS, OME-TIFF, TIFF, PNG, and JPEG (Fig.~\ref{fig:fig5}b). The interactive whole-slide viewer uses a high-performance rendering engine that streams image tiles dynamically, enabling smooth navigation across gigapixel tissue sections with real-time overlay of segmentation masks.

\begin{center}
    \includegraphics[width=\textwidth]{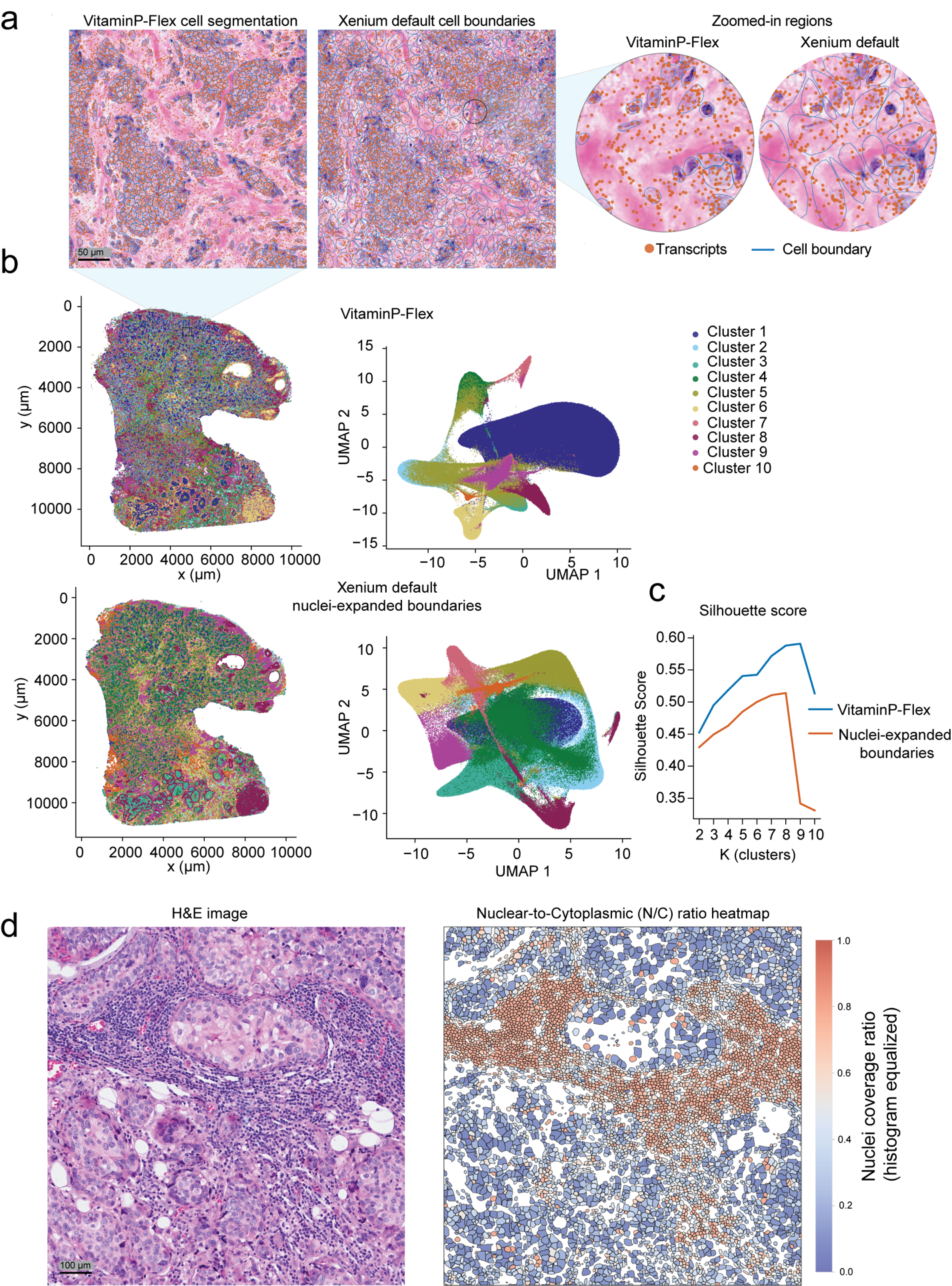}
\end{center}

\captionof{figure}{\textbf{Applications of H\&E-driven whole-cell segmentation.} \textbf{a,} Comparison of cell segmentation methods and corresponding zoomed regions. VitaminP-Flex segmentation derived from H\&E (left) is compared with Xenium default segmentation (middle), which expands nuclear boundaries from a DAPI morphology channel. Zoomed regions (right) highlight differences in boundary delineation and transcript assignment, showing that VitaminP-Flex more accurately follows histological cell contours while reducing overlap between neighboring cells. \textbf{b,} Spatial distributions (left) and UMAP projections (right; $K = 10$ clusters) demonstrate improved cluster separation and structural coherence with VitaminP-Flex segmentation. \textbf{c,} Quantitative evaluation of clustering performance across $K = 2\text{--}10$ clusters. \textbf{d,} Whole-cell morphometric analysis enabled by H\&E-driven segmentation. Nuclear-to-cytoplasmic (N/C) ratios are computed from predicted nuclear and cytoplasmic boundaries.}
\label{fig:fig4}

\vspace{1em}
AI segmentation inference is executed asynchronously via a dedicated inference service, which can be deployed locally or on GPU-accelerated systems. To handle large-scale analysis efficiently, VitaminPScope employs a tile-based inference pipeline that processes gigapixel images in a streaming manner while preserving spatial context. This design enables accurate segmentation of nuclear and whole-cell contours with seamless real-time visualization. In addition to visualization, the platform supports downstream morphometric analysis by exporting cell-level results in GeoJSON, JSON, and tabular formats, enabling integration with external spatial analysis pipelines. By combining scalable AI inference, interactive visualization, and quantitative analysis, VitaminPScope provides a practical and user-friendly solution for incorporating VitaminP segmentation models into routine histology workflows (Fig.~\ref{fig:fig5}c--e).

\section*{Discussion}

Whole-cell segmentation from H\&E images remains underexplored due to limited cytoplasmic contrast, which obscures cell boundaries in routine histology. Here, we present VitaminP as a multimodal framework that leverages complementary information from paired H\&E--mIF images during training. Rather than relying solely on appearance-based cues, VitaminP learns to infer cell boundaries by aligning morphological structure in H\&E with molecularly defined boundaries in mIF, thereby improving boundary representation. The strong cross-modal agreement between H\&E-derived predictions and mIF-based segmentation supports the fidelity of these representations, with improvements retained when applying the model to H\&E alone at inference, without requiring mIF inputs. The enhanced boundary delineation enables downstream applications, including improved spatial transcriptomic analysis and cell-type-specific morphometric quantification. To facilitate practical deployment, we further developed VitaminPScope, an open-source platform for scalable whole-slide inference, visualization, and analysis in research pathology workflows.

\begin{center}
    \includegraphics[width=\textwidth]{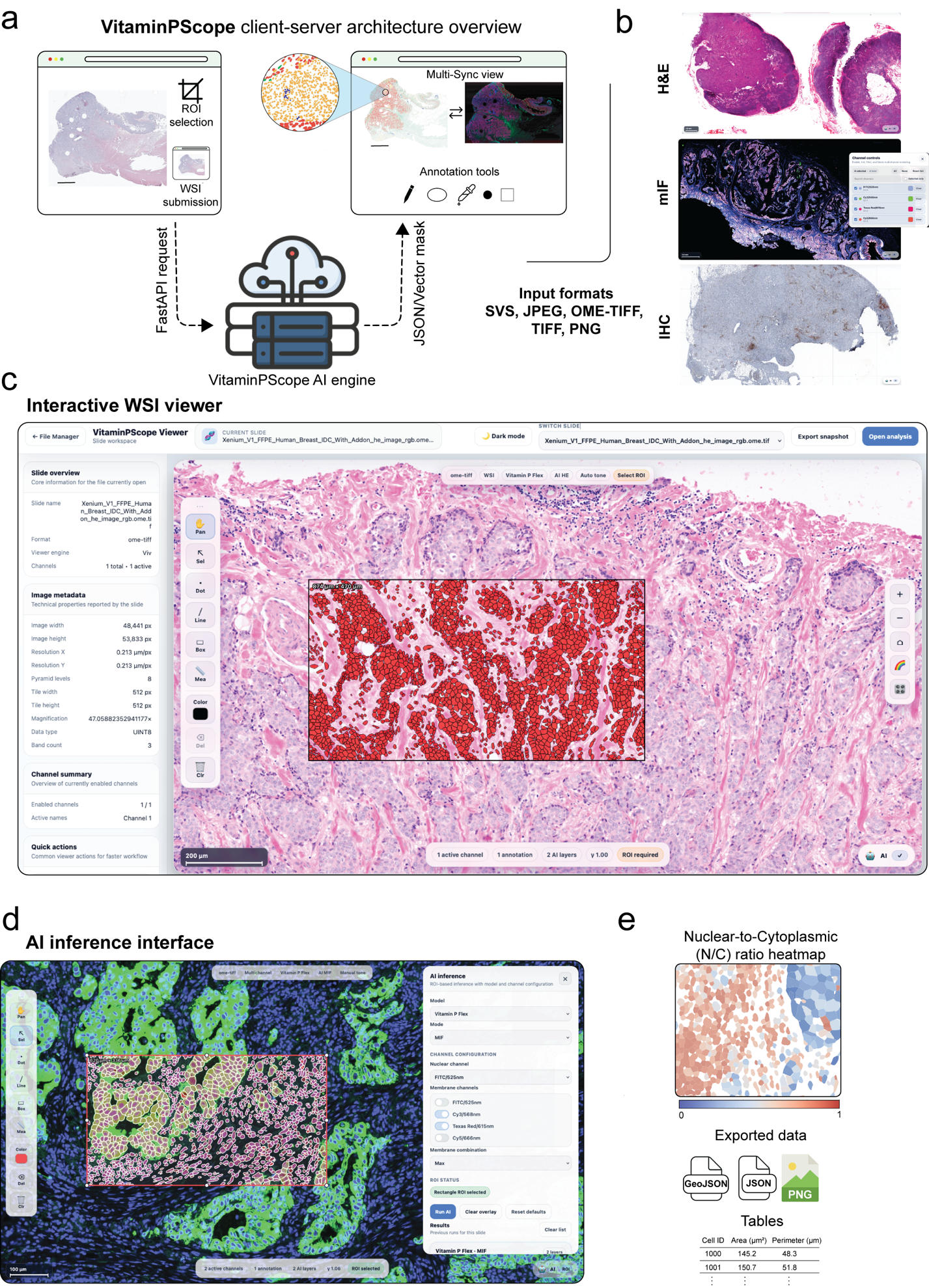}
\end{center}

\captionof{figure}{\textbf{VitaminPScope platform for interactive whole-slide image analysis and quantitative pathology outputs.} \textbf{a,} Client--server architecture of VitaminPScope. Users upload or mount datasets and select ROIs through a browser-based interface. Inference requests are sent via API to the backend VitaminPScope AI engine, which performs segmentation and returns vectorized masks for real-time visualization and downstream analysis. \textbf{b,} Supported pathology imaging modalities and file formats, including H\&E, mIF, and IHC, as well as common file formats such as SVS, OME-TIFF, TIFF, JPEG, and PNG. \textbf{c,} Interactive web-based viewer for seamless exploring of WSIs with overlaid AI segmentation and ROI-based analysis. \textbf{d,} AI inference interface for configuring segmentation tasks on multiplex imaging data. Users can select specific channels and launch inference within selected ROIs, enabling automated segmentation of cell and nuclear across large tissue regions. \textbf{e,} Quantitative outputs generated from AI-derived cell segmentation. Examples include nuclear-to-cytoplasmic (N/C) ratio heatmaps and exportable cell-level measurements in tabular format for downstream spatial and morphometric analysis.}
\label{fig:fig5}

\vspace{1em}
A central challenge in whole-cell segmentation is the scarcity of reliable annotations, particularly for H\&E images, where cytoplasmic borders are frequently ambiguous. We addressed this through an AI--human collaboration workflow, enabling efficient generation of large-scale paired annotations from ORION and 10x Xenium slides while maintaining expert control over annotation fidelity. This is important because standard public benchmarks do not adequately capture the difficulty of whole-cell delineation in H\&E images. Our results therefore support both the scalability of the curation framework and the value of cross-modal supervision in challenging settings where precise cell boundaries are critical.
\vspace{1em}

Precise delineation of cell boundaries directly influences downstream biological interpretation in spatial omics workflows. Improved cellular boundary definition enhances the fidelity of transcript-to-cell assignment and preserves biologically meaningful structure in gene expression space. Consistent with this, segmentation derived from H\&E using VitaminP yielded more coherent and better-separated transcriptional clusters compared to nuclei-expansion strategies. Although VitaminP-derived segmentations resulted in modest changes in median transcripts per cell ($-24\%$ to $+4\%$ across samples) relative to nuclei-based expansion, downstream clustering analyses demonstrated improved structure, likely reflecting increased specificity of transcript assignment rather than reduced sensitivity. A recent study reported that fewer than 5\% of nuclei are intact in 5\,$\mu$m tissue sections~\cite{yapp_highly_2025}, suggesting that cytoplasm may be captured without a corresponding nucleus. As nuclei-based expansion methods rely on nuclei as anchors for cell segmentation, transcripts from cells without captured nuclei may therefore be misassigned to neighboring nucleated cells.
\vspace{1em}

To facilitate translation into real-world workflows, we developed VitaminPScope, an open-source platform for scalable whole-slide inference and interactive analysis of cell-level segmentations. By integrating VitaminP within a modular client--server architecture, the platform enables efficient processing of gigapixel images through tile-based streaming while supporting interactive visualization. This design provides a unified interface for segmentation, ROI exploration, and quantitative cell-level analysis, lowering the barrier to adoption in computational and clinical pathology settings.
\vspace{1em}

Despite these advances, several limitations warrant consideration. First, molecular supervision from mIF is constrained by marker selection; the limited set of membrane-associated markers may not capture the full diversity of cell boundary phenotypes. Second, evaluation of whole-cell segmentation remains challenging due to incomplete ground truth, and conservative annotations may underestimate performance by penalizing correctly predicted but unlabeled cells. Finally, the framework operates on two-dimensional sections and depends on spatially aligned paired data; registration errors and acquisition variability may introduce noise. Extending the framework to three-dimensional or multimodal data will be important for improving robustness and biological interpretability.
\vspace{1em}

Collectively, these results establish cross-modal learning from paired H\&E and mIF images as a practical and scalable strategy for accurate whole-cell segmentation from routine histology alone. By linking morphological features with molecularly defined boundaries during training, VitaminP enables robust boundary delineation that generalizes across diverse tissues and imaging modalities. Coupled with the scalable inference capabilities of VitaminPScope, VitaminP provides both a methodological framework and a practical foundation for advancing cell-level computational pathology toward large-scale, multimodal, and translational applications.

\printbibliography

\section*{Methods}

\subsection*{Dataset construction}

To enable cross-modal supervision while preserving morphological diversity, we constructed a multi-source training dataset that combined spatially paired H\&E--mIF images with large-scale public nuclear and whole-cell segmentation datasets. The paired data consisted primarily of the ORION colorectal cancer (CRC) cohort~\cite{lin_highly_2018} and the 10x Xenium public dataset~\cite{10xgenomics_datasets}, which together provide explicit molecular boundary supervision for transferring cytoplasmic information to H\&E.
\vspace{1em}

The ORION dataset included 41 whole-slide image (WSI) pairs, each containing spatially aligned brightfield H\&E and mIF images. The mIF panels comprised 19 fluorescence channels. For segmentation supervision, we selected structurally informative mIF channels: DAPI for nuclear boundary delineation, together with the membrane-associated markers E-cadherin and CD45 to delineate epithelial and immune cell boundaries, respectively. To improve robustness across heterogeneous tumor microenvironments, we constructed a composite membrane representation by integrating E-cadherin and CD45 signals, enabling unified whole-cell supervision across epithelial nests and immune-rich stromal regions. All ORION images were acquired at a resolution of 0.325~$\mu$m per pixel (MPP), and H\&E--mIF spatial correspondence was preserved at the slide level before patch tiling.
\vspace{1em}

To complement the paired datasets and promote robustness to staining and morphological variability, we incorporated 12 additional publicly available segmentation datasets spanning diverse organs, tumor types, and staining protocols, resulting in a total of 14 datasets covering 34 cancer types. These included nucleus-focused benchmarks such as PanNuke~\cite{gamper_pannuke_2020}, Lizard~\cite{graham_lizard_2021}, MoNuSeg~\cite{kumar_multi-organ_2020}, TNBC~\cite{naylor_segmentation_2019}, NuInsSeg~\cite{mahbod_nuinsseg_2024}, CryoNuSeg~\cite{mahbod_cryonuseg_2021}, BC (breast carcinoma)~\cite{amgad_structured_2019}, CoNSeP~\cite{graham_hover-net_2019}, MoNuSAC~\cite{verma_monusac2020_2021}, Kumar~\cite{kumar_dataset_2017}, and CPM-17~\cite{vu_methods_2019}, as well as datasets with explicit whole-cell annotations such as TissueNet~\cite{greenwald_whole-cell_2022}. Together, these datasets contributed more than seven million annotated instances across epithelial, mesenchymal, immune, and mixed tumor contexts. The inclusion of data acquired from different scanners, staining protocols, and tissue preparation conditions was intended to mitigate overfitting to stain- or site-specific artifacts and to promote learning of broadly applicable histomorphological features.
\vspace{1em}

High-fidelity supervision was generated using an AI--human collaboration workflow. Initial nuclear and whole-cell masks were generated using modality-specialized segmentation models optimized separately for H\&E and mIF inputs. For paired H\&E--mIF datasets, whole-cell masks were derived from mIF images and transferred to the corresponding H\&E images, providing cross-modal supervision for model training. Pathologists then reviewed these predictions and selected high-confidence regions of interest (ROIs) with reliable boundary delineations for subsequent tiling into fixed-size patches. This curated paired subset yielded millions of high-quality nuclear and whole-cell annotations from the ORION and Xenium datasets, forming the core cross-modal supervision signal.
\vspace{1em}

All datasets were partitioned using a stratified cross-validation strategy that balanced tissue types, acquisition sources, and modalities to minimize dataset- and modality-specific bias.

\subsection*{VitaminP architecture}

VitaminP is a dense, single-pass instance segmentation framework designed for whole-slide scalability, in which runtime scales linearly with image area rather than with the number of cells. Across all variants, the architecture follows a transformer--UNet hybrid design: a hierarchical vision transformer (ViT) encoder produces multi-resolution feature maps that are decoded through UNet-style~\cite{ronneberger_u-net_2015} upsampling blocks with skip connections to generate dense prediction maps. The transformer encoder was initialized from self-supervised pretrained weights. The primary methodological contribution lies in the cross-modal learning strategy and architectural design that enable paired modality fusion and efficient dense inference, rather than the specific pretraining scheme.
\vspace{1em}

VitaminP-Dual implements a paired-modality architecture that separates modality-specific representation learning in early layers while enforcing shared semantic structure in deeper layers. H\&E (3-channel) and mIF-derived inputs (2-channel: DAPI and membrane composite) are processed by two independent encoders to extract low- and mid-level features. Features from the first transformer stage are fused by channel-wise concatenation followed by a learnable $1\times1$ projection (``fusion bottleneck''), producing a joint representation that integrates complementary cues from both modalities. Higher-level features are then computed using a shared deeper encoder, providing a unified latent space for decoding under paired supervision.
\vspace{1em}

To support simultaneous prediction across modalities, VitaminP-Dual incorporates four dedicated decoder branches: H\&E nucleus, H\&E whole-cell, mIF nucleus, and mIF whole-cell outputs. Each branch follows a UNet-style decoding path with bilinear upsampling and multi-scale skip connections. The final high-resolution decoding stage concatenates early features from both modalities, enabling fine-grained boundary refinement that leverages paired signals even for H\&E-only predictions. Each decoder predicts a binary segmentation probability map together with a two-channel horizontal--vertical (HV) offset field, facilitating instance separation without proposal generation or non-maximum suppression. Deep encoder features are regularized using spatial dropout to stabilize training of high-capacity backbones.
\vspace{1em}

VitaminP-Syn adopts the same dual-encoder, fused-latent, multi-branch decoder design as VitaminP-Dual, but replaces the real mIF input with a synthetic two-channel representation generated from H\&E using a Pix2Pix-style~\cite{isola2017} conditional generative adversarial network (GAN). This synthetic pathway preserves the benefits of paired-modality supervision while enabling inference in H\&E-only settings. Both training and inference use two inputs (real H\&E and a synthetic mIF proxy), allowing the model to maintain a consistent fusion-and-decode mechanism across paired and unpaired scenarios, with identical segmentation heads and post-processing.
\vspace{1em}

VitaminP-Flex is a streamlined generalist variant optimized for high-throughput inference across heterogeneous inputs. It uses a single shared encoder followed by an atrous spatial pyramid pooling (ASPP) module~\cite{chen_rethinking_2017} applied to the deepest feature map, capturing multi-scale context to enhance robustness to variations in cell size, density, and tissue architecture. In the decoder, conventional normalization and activation layers are replaced with GroupNorm~\cite{wu_group_2018} and GELU~\cite{hendrycks_gaussian_2016} to ensure stable training under distributed batch sizes, while lightweight squeeze-and-excitation channel attention~\cite{hu_squeeze-and-excitation_2017} is integrated into convolutional decoder blocks to recalibrate features for boundary-sensitive prediction.
\vspace{1em}

Each prediction head incorporates a CoordConv refinement stage~\cite{liu_intriguing_2018} that appends normalized $x$--$y$ coordinate channels before generating the segmentation map and HV offset field, explicitly encoding spatial priors to improve instance separation and reduce boundary artifacts. Like VitaminP-Dual and VitaminP-Syn, VitaminP-Flex produces both nuclear and whole-cell outputs in a single forward pass, avoiding iterative prompting or object-wise decoding and ensuring constant-time inference independent of cell count.

\subsection*{VitaminP training}

VitaminP was trained using a multi-task objective that combined (i) binary segmentation of nuclear and whole-cell masks with (ii) horizontal--vertical (HV) regression maps for instance separation. For segmentation, we used a hybrid loss consisting of Dice loss~\cite{milletari_v-net_2016} and focal binary cross-entropy~\cite{lin_focal_2017} ($\alpha = 1$, $\gamma = 2$). For HV prediction, we adopted a Hover-Net--style loss that included mean squared error (MSE) on the HV vectors and a mean squared gradient error (MSGE) term, computed using $5 \times 5$ Sobel-like derivative kernels; the gradient loss was masked to foreground regions using the corresponding binary mask.
\vspace{1em}

In the dual-modality models, VitaminP-Dual and VitaminP-Syn, losses were computed independently across four decoder heads (H\&E nuclei, H\&E whole-cells, mIF nuclei, mIF whole-cells). HV losses were weighted $2\times$ relative to the segmentation loss, and the final loss was averaged across all heads.
\vspace{1em}

For the flexible model (VitaminP-Flex), each training sample was routed to the corresponding modality-specific decoder. Nuclear HV predictions were weighted $4\times$ and whole-cell HV predictions $2\times$ relative to segmentation; whole-cell losses were omitted for datasets lacking cell annotations.
\vspace{1em}

All models were optimized using AdamW~\cite{loshchilov_decoupled_2017} (learning rate $= 1\times10^{-4}$, weight decay $= 1\times10^{-2}$) with gradient norm clipping (max norm $= 0.5$) and a \texttt{ReduceLROnPlateau} scheduler (factor $= 0.8$, patience $= 5$) monitored on validation loss. Training employed extensive data augmentation, including random flips and rotations, scaling and random resizing (to promote magnification invariance), stain and color perturbations, Gaussian blur, additive noise, cutout, and percentile-based intensity normalization.

\subsection*{VitaminP inference}

At inference, VitaminP enables efficient whole-slide image (WSI) processing through streaming, tile-wise execution rather than object-by-object prompting. Full WSIs are not loaded into memory; instead, slides are accessed via a reader interface and processed as a grid of $512 \times 512$ tiles with 64-pixel overlap. Optionally, tiles are pre-filtered using a modality-aware tissue detector (with differentiated logic for brightfield and fluorescence), and the resulting tissue mask is morphologically dilated to include boundary regions and prevent truncation artifacts.
\vspace{1em}

To ensure resolution consistency across scanners, each slide is normalized to a target resolution. When available, the microns-per-pixel (MPP) value is extracted from OpenSlide/OME-TIFF metadata~\cite{goode_openslide_2013} and used to compute a virtual scaling factor relative to the training MPP. Tiles are extracted on demand, resized to $512 \times 512$, normalized (including H\&E-specific preprocessing), and processed in GPU batches (default batch size = 8) to maximize throughput. This design ensures that wall-clock time scales primarily with tissue area and batching efficiency rather than the number of detected objects.
\vspace{1em}

VitaminP supports both single- and dual-modality inference. For VitaminP-Dual, multiplexed features can be provided as a registered two-channel mIF image (nuclear and membrane channels). When mIF is unavailable, synthetic mIF is optionally generated per tile using a Pix2Pix-based generator that translates the H\&E tile into a two-channel representation prior to segmentation. This synthesis step is seamlessly integrated into the same per-tile pipeline and is activated only when synthetic mIF generation is enabled.
\vspace{1em}

Instance extraction is performed per tile to avoid full-slide stitching. For nuclear branches, instances are recovered from the predicted segmentation and HV maps using a Hover-Net--style watershed algorithm applied to an energy landscape derived from the HV fields. For whole-cell branches (when both nuclear and cell outputs are requested), a nuclear-constrained HV watershed is used: nuclear instance labels act as markers to enforce a one-cell-per-nucleus prior, while HV-derived gradients define boundary walls. Additional post-processing includes recovery of instances lacking associated nuclei, in which cell blobs without nuclei are detected and seeded via distance-transform maxima, enabling recovery of anucleate or weak-nucleus regions while maintaining consistent geometry.
\vspace{1em}

Following local instance extraction, contours, bounding boxes, and centroids are transformed from tile-local to global slide coordinates. To eliminate duplicates in overlapping regions, only instances whose centroids fall within the central ``core'' region of each tile are retained, defined by excluding a margin equal to half of the overlap from each tile boundary. For outermost tiles, margins are relaxed to preserve boundary objects. Remaining duplicates are removed using intersection-over-union (IoU)-based polygon deduplication. Results are exported in standard formats (JSON, GeoJSON, Parquet) with optional visual overlays.

\subsection*{Benchmarking and evaluation}

\subsection*{Public multi-dataset benchmarking}

We benchmarked VitaminP-Flex as a generalist model across 13 publicly available datasets spanning a wide range of tissues, staining protocols, and acquisition modalities (Fig.~\ref{fig:fig2}a; Extended Data Fig.~\ref{fig:ext_fig3}a), including H\&E, chromogenic IHC, and fluorescence images. We compared its performance against four established methods representing different segmentation paradigms: CellSAM (prompt-based), Cellpose-SAM (combining SAM-based feature encoding with Cellpose-style flow-based segmentation), Hover-Net (convolutional), and CellViT++ (transformer-based).
\vspace{1em}

For Hover-Net and CellViT++, we used the official pretrained weights released by the authors. Both models were originally trained on PanNuke, although the precise train/test partition used to generate the released weights is not fully documented. CellViT++ reports using up to 95\% of available data for its final model, but the split configuration remains unspecified. For CellSAM and Cellpose-SAM, the training data reported in their respective publications include several of the public datasets included in our evaluation benchmark. As a result, some evaluation datasets may overlap with those used during pretraining or fine-tuning.
\vspace{1em}

All models were evaluated using their publicly released weights without additional retraining or dataset-specific fine-tuning. The primary evaluation metric for instance segmentation was Panoptic Quality (PQ)~\cite{kirillov_panoptic_2018}, which jointly measures detection and segmentation performance. PQ is defined as follows:
\begin{equation}
\mathrm{PQ} = \frac{\sum_{(p,g)\in \mathrm{TP}} \mathrm{IoU}(p,g)}{|\mathrm{TP}| + \tfrac{1}{2}|\mathrm{FP}| + \tfrac{1}{2}|\mathrm{FN}|}
\end{equation}
where predicted instances $p$ and ground truth instances $g$ were matched using an intersection-over-union (IoU) threshold of 0.5. True positives (TP), false positives (FP), and false negatives (FN) were determined based on this matching criterion. PQ penalizes both missed detections and inaccurate boundary delineations and served as the primary ranking metric.
\vspace{1em}

To further evaluate robustness and generalization, we quantified the frequency with which each model ranked in the Top-1, Top-2, and Top-3 positions according to PQ scores across the benchmarked datasets, thereby assessing performance stability across diverse conditions.

\subsection*{Pixel-level and rare cancer validation}
To assess boundary fidelity independently of instance matching, we computed the Dice coefficient:
\begin{equation}
\mathrm{Dice}(P,G) = \frac{2|P \cap G|}{|P| + |G|}
\end{equation}
where $P$ denotes the predicted binary mask and $G$ denotes the ground truth mask. Dice was primarily used to quantify cross-modal agreement. The rare cancer dataset comprised 24 image tiles exhibiting morphologies absent from the training data. Nuclear annotations were exhaustive, whereas whole-cell annotations were restricted to high-confidence regions. As some cytoplasmic regions remained partially annotated, the resulting evaluation metrics represent conservative (lower-bound) estimates of segmentation accuracy.

\subsection*{Zero-shot external evaluation}

To assess out-of-distribution generalization, we conducted zero-shot evaluation on three external datasets: CPM-15, PanopTILs~\cite{liu_panoptic_2024}, and DSB2018. VitaminP model weights were frozen after the multi-dataset training phase; no fine-tuning, parameter adaptation, or dataset-specific normalization was applied to these external cohorts.
\vspace{1em}

Comparator models were evaluated using their publicly released pretrained weights. Notably, CellSAM and Cellpose-SAM report pretraining on datasets that include DSB2018 or closely related fluorescence microscopy data, indicating that their performance on DSB2018 may reflect partial overlap with training data. Hover-Net and CellViT++ were evaluated using their official pretrained checkpoints, both of which were trained primarily on PanNuke. No additional retraining or dataset-specific tuning was applied. PQ served as the primary evaluation metric to maintain consistency with the benchmarking framework described above.

\subsection*{Cross-modal consistency analysis}

To evaluate geometric concordance across modalities, we compared whole-cell segmentations derived from H\&E images with those obtained from mIF on paired tiles. Dice coefficients were computed between H\&E-predicted masks and mIF-derived segmentations (produced by external mIF segmentation pipelines), providing a measure of cross-modal consistency that is independent of limitations in manual annotations.

\subsection*{Synthetic mIF evaluation for VitaminP-Syn}

To quantify the effectiveness of synthetic cross-modal supervision, we evaluated VitaminP-Syn by comparing segmentation performance when using synthetic mIF representations versus paired real mIF inputs. Under identical evaluation conditions, the synthetic pathway achieved near parity with VitaminP-Dual, retaining approximately 99.9\% of the dual-modality performance (Fig.~\ref{fig:fig2}a; Extended Data Fig.~\ref{fig:ext_fig1}a--c). Quantitative similarity metrics, together with representative visual examples, confirmed the faithful preservation of nuclear and membrane signal structure in the synthetic mIF representations (Extended Data Fig.~\ref{fig:ext_fig1}).

\subsection*{Whole-slide scalability benchmarking}

To evaluate deployment efficiency, we benchmarked inference scalability across five metrics: single-tile latency (milliseconds per $512 \times 512$ px tile), whole-slide processing time (minutes per $15{,}000 \times 15{,}000$ px image), peak GPU memory consumption (GB), segmentation throughput (cell instances per second), and area processing rate (megapixels per minute; MPx/min). All experiments were conducted on a single NVIDIA H100 GPU (80 GB memory), with GPU model, driver version, and batch size fixed across models to ensure consistent hardware conditions. Tile-level latency was measured over ten inference runs following three warm-up iterations, with GPU synchronization performed before and after each run for precise timing. Peak GPU memory consumption was recorded using GPU memory allocation statistics at the start of each run. Whole-slide inference was evaluated on a representative $15{,}000 \times 15{,}000$ px H\&E image, processed as overlapping $512 \times 512$ px tiles (64 px overlap). End-to-end wall-clock time was measured from initial slide ingestion through final metric computation and mask export, capturing the complete inference pipeline rather than model forward-pass time alone. Normalized performance ratios reported in the main text are expressed relative to CellSAM single-task inference (set as the $1\times$ baseline). Instance counts were computed by enumerating unique non-background labels in the output segmentation masks, and segmentation throughput was calculated as the total number of detected instances divided by total elapsed time.

\subsection*{Hyperparameters}

All models were trained end-to-end using the AdamW optimizer with an initial learning rate of $1\times10^{-4}$ and a uniform weight decay of $1\times10^{-2}$ applied to all parameters. We did not employ layer-wise learning rate decay or staged backbone freezing. The learning rate was scheduled using \texttt{ReduceLROnPlateau}, which reduced the learning rate by a factor of 0.8 after five consecutive epochs without improvement in validation loss. The model checkpoint was selected based on the highest validation Dice score rather than the lowest validation loss.
\vspace{1em}

Training was performed with a batch size of 4 per iteration, as defined in the data loader configuration, with mixed multi-dataset batches sampled from ORION, 10x Xenium, TissueNet, PanNuke, Lizard, MoNuSeg, TNBC, NuInsSeg, CryoNuSeg, BC, CoNSeP, Kumar, CPM-17, and MoNuSAC. Gradient clipping (maximum norm = 0.5) was applied to stabilize optimization, particularly given the inclusion of HV regression heads.
\vspace{1em}

The segmentation objective combined Dice and Focal loss for binary mask prediction with a Hover-Net--style HV regression loss. The HV branch incorporated both mean squared error (MSE) and mean squared gradient error (MSGE) terms to penalize gradient inconsistencies in the horizontal and vertical distance maps, thereby improving instance separation. In the Dual configuration, HV losses were weighted twice as strongly as segmentation losses. In Flex and baseline configurations, nuclear HV losses received a weight of 4.0 and whole-cell HV losses a weight of 2.0, reflecting the stricter structural constraints required for accurate nuclear boundary delineation. HV maps were generated on the fly using the PanNuke gradient-normalized formulation, which computes instance-wise horizontal and vertical distance maps with center-of-mass normalization and direction-specific scaling (positive and negative). This approach improves boundary precision and reduces merging artifacts relative to centroid-only normalization.
\vspace{1em}

Extensive data augmentation was applied to improve robustness to magnification and staining variability. Geometric augmentations included random horizontal/vertical flips, 90$^\circ$/180$^\circ$/270$^\circ$ rotations, isotropic scaling (0.5--2.0$\times$), and random resizing. Photometric perturbations included brightness/contrast adjustments, hue/saturation shifts for stain perturbation, Gaussian blur, additive noise, and random cutout. All augmentations were disabled during validation and testing. Experiments used a consistent train/validation/test split defined at the sample level to prevent patch-level leakage. Complete configuration details specifying dataset composition, augmentation probabilities, HV generation methods, and data loader settings are provided in the accompanying code repository to ensure reproducibility.

\subsection*{VitaminPScope system architecture}

VitaminPScope is implemented as a modular client--server platform designed for scalable deployment of deep learning--based segmentation models on whole-slide pathology images. The system comprises three core components: (1) a web-based frontend for interactive visualization and result exploration; (2) an application backend for data management, job orchestration, and user session handling; and (3) a dedicated AI inference service for model execution.
\vspace{1em}

Segmentation inference was performed asynchronously by the AI service, which exposed RESTful endpoints via a FastAPI interface. Upon submission of an inference request, the backend coordinated model execution and returned vectorized segmentation outputs encoded as GeoJSON objects, representing individual cell boundaries along with associated metadata. The inference service can be deployed locally or on remote compute infrastructure, including GPU-enabled workstations or cloud environments, enabling flexible use of available hardware resources. These outputs were visualized directly within the web interface, allowing real-time inspection of predicted nuclear and whole-cell segmentations across large tissue regions.
\vspace{1em}

To ensure scalability for gigapixel pathology images, VitaminPScope employed a streaming, tile-based inference pipeline. WSIs were accessed through format-specific readers (e.g., OpenSlide or OME-TIFF) and decomposed into overlapping tiles that were processed sequentially while preserving global spatial coordinates. Instance segmentation results were reconstructed from per-tile predictions and merged into slide-level outputs using boundary-aware overlap resolution. This design ensured that inference time scaled primarily with image area rather than cell count, enabling efficient analysis of WSIs containing millions of cells.

\subsection*{AI inference service}

The AI inference service was implemented using FastAPI and exposed RESTful endpoints for segmentation requests. Each incoming inference request included model configuration parameters such as the selected VitaminP variant (VitaminP-Dual, VitaminP-Syn, or VitaminP-Flex), checkpoint weights, tile size, tile overlap, and target spatial resolution. Upon receiving a request, the service dynamically loaded the requested model checkpoint (with caching to prevent redundant loading) and executed whole-slide inference through the VitaminP inference pipeline. The service supported multiple model variants through a unified adapter interface that manages model loading, parameter configuration, and execution logic. Model inference was executed on GPU or CPU depending on the deployment configuration.

\subsection*{Whole-slide inference pipeline}

Whole-slide segmentation employed a streaming, tile-based inference strategy to prevent loading entire gigapixel images into memory. Slides were accessed using a multi-format image loader supporting OpenSlide-compatible formats (e.g., SVS), OME-TIFF, and standard image formats. The loader exposed a unified tile-streaming interface, enabling on-demand retrieval of arbitrary image regions.
\vspace{1em}

Each slide was decomposed into overlapping tiles of fixed size (typically $512 \times 512$ pixels), with configurable overlap to mitigate boundary artifacts. Each tile was preprocessed and fed to the VitaminP model in batches to optimize GPU utilization. Model outputs consisted of segmentation probability maps and horizontal--vertical (HV) offset fields for instance separation.
\vspace{1em}

Instance extraction was performed following tile inference to minimize peak memory usage. For nuclear segmentation, instance masks were generated using a Hover-Net--style watershed algorithm applied to the predicted HV vector fields. For whole-cell segmentation, a nuclei-constrained watershed was used, in which nuclear instances served as seeds for delineating corresponding cellular boundaries. All extracted instances were then converted from tile-local coordinates into global slide coordinates.

\subsection*{Overlap resolution and instance merging}

Because tiles overlap spatially, predicted instances may occur multiple times near tile boundaries. VitaminPScope resolved these duplicates through a boundary-aware filtering strategy that retained only instances whose centroids fall within the central tile core region. Remaining duplicates were eliminated using IoU-based polygon deduplication. This approach preserved segmentation accuracy while preventing redundant detections at tile borders.

\subsection*{Output representation and visualization}

Segmentation outputs were exported as vectorized geometric objects, each containing cell contours (polygons), centroids, bounding boxes, and optional morphometric measurements (e.g., area, perimeter, circularity). These results were serialized as GeoJSON FeatureCollections, enabling seamless rendering within the web-based viewer and interoperability with downstream spatial analysis tools.

\subsection*{Web interface and visualization}

The VitaminPScope frontend was implemented using a modern JavaScript framework and integrated a high-performance whole-slide image viewer supporting multiscale tile rendering and real-time overlay of segmentation results. Users were able to interactively explore large slides, configure segmentation parameters, submit AI inference jobs, and visualize predicted nuclei and whole-cell boundaries directly overlaid on tissue images. Results could be exported in standardized formats for downstream computational analysis or integration into spatial omics workflows.

\subsection*{Data availability}

The in-house curated rare cancer images and annotations are available at \url{https://zenodo.org/records/19355453}. All other datasets used in this study are publicly available from their respective sources.

\subsection*{Code availability}

The VitaminP models and the VitaminPScope platform were implemented in Python and JavaScript and are distributed as open-source software. The platform includes a containerized deployment environment supporting GPU-accelerated inference. Source code for model training, inference pipelines, and the VitaminPScope visualization platform is publicly available at \url{https://github.com/idso-fa1-pathology/VitaminP}. Model weights are available at \url{https://huggingface.co/idso-fa1-pathology/VitaminP}.

\subsection*{Acknowledgements}

This work was supported by the Spatial Ecology \& Quantitative Pathology Image Analytical platform (SEQUOIA) through the MD Anderson Strategic Research Initiative Development Program (STRIDE) and the Patient Mosaic™ project at The University of Texas MD Anderson Cancer Center. This project was funded by Lyda Hill Philanthropies. Patient Mosaic is supported by generous philanthropic contributions from the Albert and Margaret Alkek Foundation, among others.

\subsection*{Author contributions}

Conceptualization: Y.S., Y.Y., P.C., and X.P. \\
Methodology: Y.S., P.C., and X.P. \\
Data curation: Y.S., K.P.G., E.B.T., and P.A. \\
Software and formal analysis: Y.S. and P.A. \\
Pathology validation and interpretation: K.P.G. and E.B.T. \\
Project administration: Y.Y., P.C., and X.P. \\
Supervision: Y.Y., P.C., and X.P. \\
Writing--original draft: Y.S., P.C., and X.P. \\
Writing--review \& editing: All authors.

\subsection*{Competing interests}

The authors declare no competing interests.

\clearpage
\section*{Extended Data}

\setcounter{figure}{0}
\renewcommand{\thefigure}{\arabic{figure}}
\renewcommand{\theHfigure}{ext.\arabic{figure}}

\noindent\makebox[\textwidth]{%
\includegraphics[width=\textwidth]{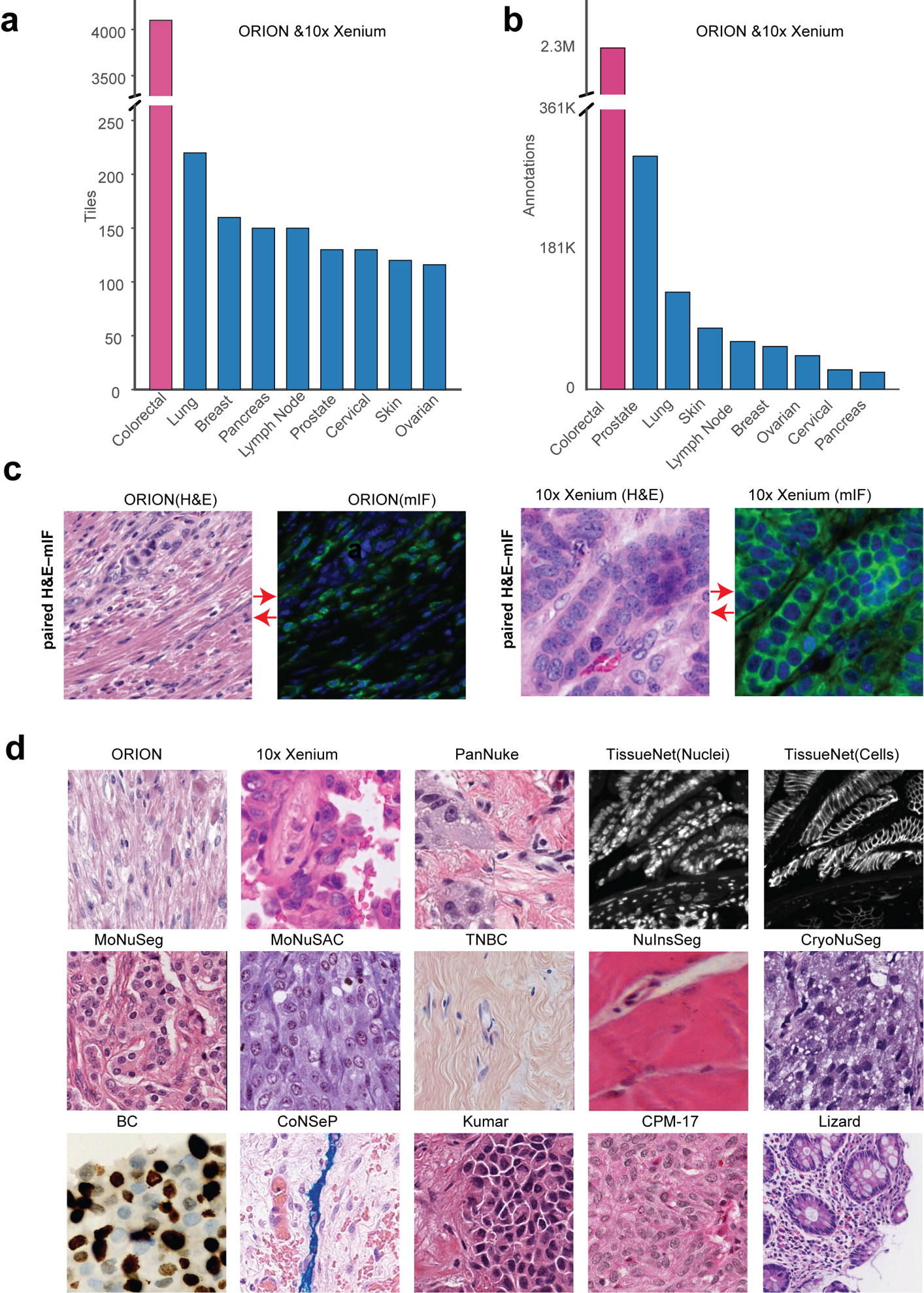}
}

\captionsetup{labelformat=empty}
\captionof{figure}{
\textbf{Extended Data Fig. 1: Dataset composition and representative examples used for training and paired cross-modal supervision.}
}
\textbf{a,} Number of image tiles curated from each dataset used for training and benchmarking, including paired datasets (ORION and 10x Xenium) and publicly available datasets spanning diverse tissues and staining modalities.
\textbf{b,} Number of instance-level annotations (nuclei and/or cells) contributed by each dataset, highlighting the scale of the training corpus.
\textbf{c,} Examples of paired H\&E and mIF image patches from the ORION and 10x Xenium datasets used for cross-modal supervision. Arrows indicate spatial correspondence between paired modalities, illustrating complementary nuclear and membrane signal information.
\textbf{d,} Representative image patches from publicly available datasets used during training and benchmarking, including ORION, 10x Xenium, PanNuke, TissueNet (nuclei and cells), MoNuSeg, MoNuSAC, TNBC, NuInsSeg, CryoNuSeg, BC, CoNSeP, Kumar, CPM-17, and Lizard, demonstrating the diversity of tissues, staining protocols, and acquisition modalities included in the study.

\label{fig:ext_fig1}
\captionsetup{labelformat=default}

\noindent\makebox[\textwidth]{%
\includegraphics[width=\textwidth]{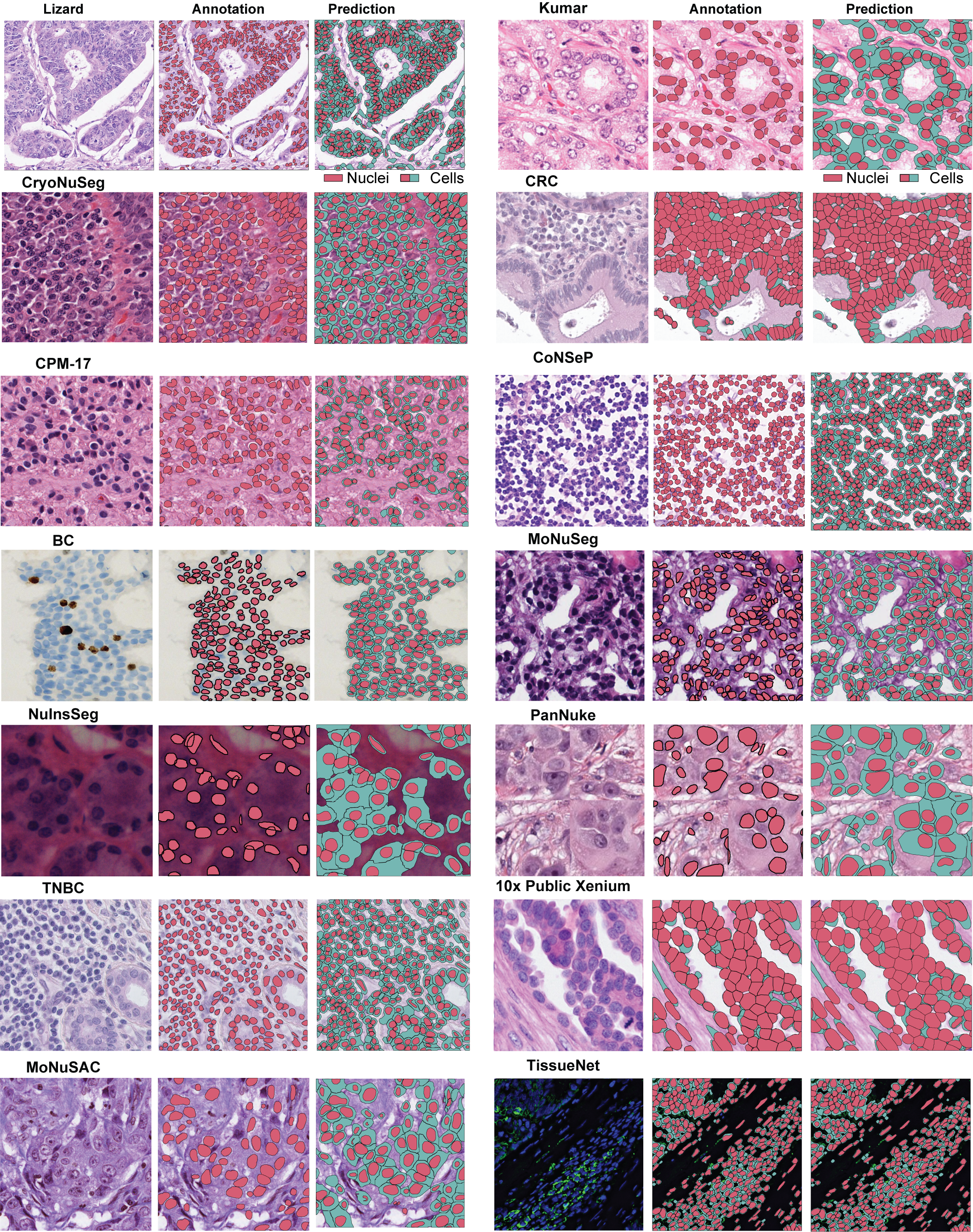}
}

\captionsetup{labelformat=empty}
\captionof{figure}{
\textbf{Extended Data Fig. 2: Qualitative comparison of whole-cell and nuclear segmentation across diverse datasets.}
Representative examples from multiple public datasets spanning varied tissue types, staining conditions, and imaging modalities, including Lizard, Kumar, CryoNuSeg, CPM-17, CoNSeP, MoNuSeg, PanNuke, TNBC, BC, NuInsSeg, MoNuSAC, CRC, and 10x Xenium, as well as TissueNet.
For each dataset, input images (left), ground-truth annotations (middle), and VitaminP predictions (right) are shown.
Nuclei are outlined in red and whole-cell regions are shown in cyan.
Across datasets, VitaminP accurately delineates nuclear and cellular boundaries despite substantial variation in morphology, staining quality, cell density, and imaging modality, demonstrating robust generalization.
}
\label{fig:ext_fig2}
\captionsetup{labelformat=default}

\noindent\makebox[\textwidth]{%
\includegraphics[width=\textwidth]{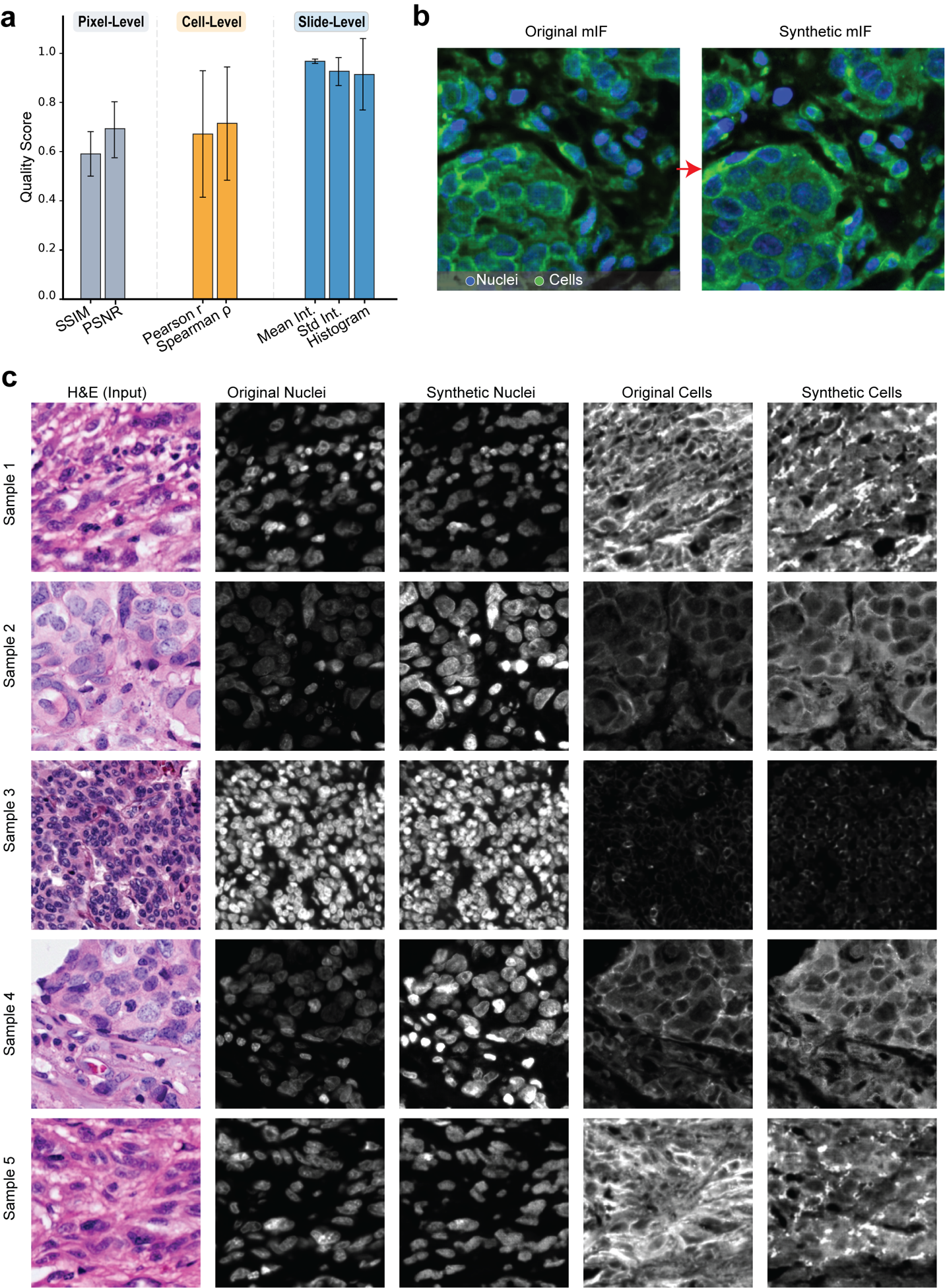}
}

\captionsetup{labelformat=empty}
\captionof{figure}{
\textbf{Extended Data Fig. 3: Quantitative and qualitative evaluation of synthetic mIF representations.}
\textbf{a,} Quantitative assessment of synthetic mIF image quality relative to paired original mIF images. Pixel-level similarity metrics include structural similarity index (SSIM) and peak signal-to-noise ratio (PSNR). Cell-level correspondence is evaluated using Pearson and Spearman correlation coefficients computed on nuclear and membrane channels. Slide-level consistency is summarized using mean intensity distribution alignment (mean intensity difference and histogram similarity). Error bars represent s.d. across evaluation tiles.
\textbf{b,} Representative example of paired original and synthetic mIF images. Synthetic mIF preserves nuclear (DAPI) and membrane-associated signal structure, maintaining spatial coherence with the corresponding original mIF image.
\textbf{c,} Qualitative comparison across five representative samples. Columns show H\&E input, original mIF nuclei channel, synthetic mIF nuclei channel, original mIF membrane/cell channel, and synthetic mIF membrane/cell channel. Across diverse morphologies and staining contexts, synthetic mIF recovers salient nuclear morphology and membrane topology while preserving global spatial organization.
}
\label{fig:ext_fig3}
\captionsetup{labelformat=default}

\noindent\makebox[\textwidth]{%
\includegraphics[width=\textwidth]{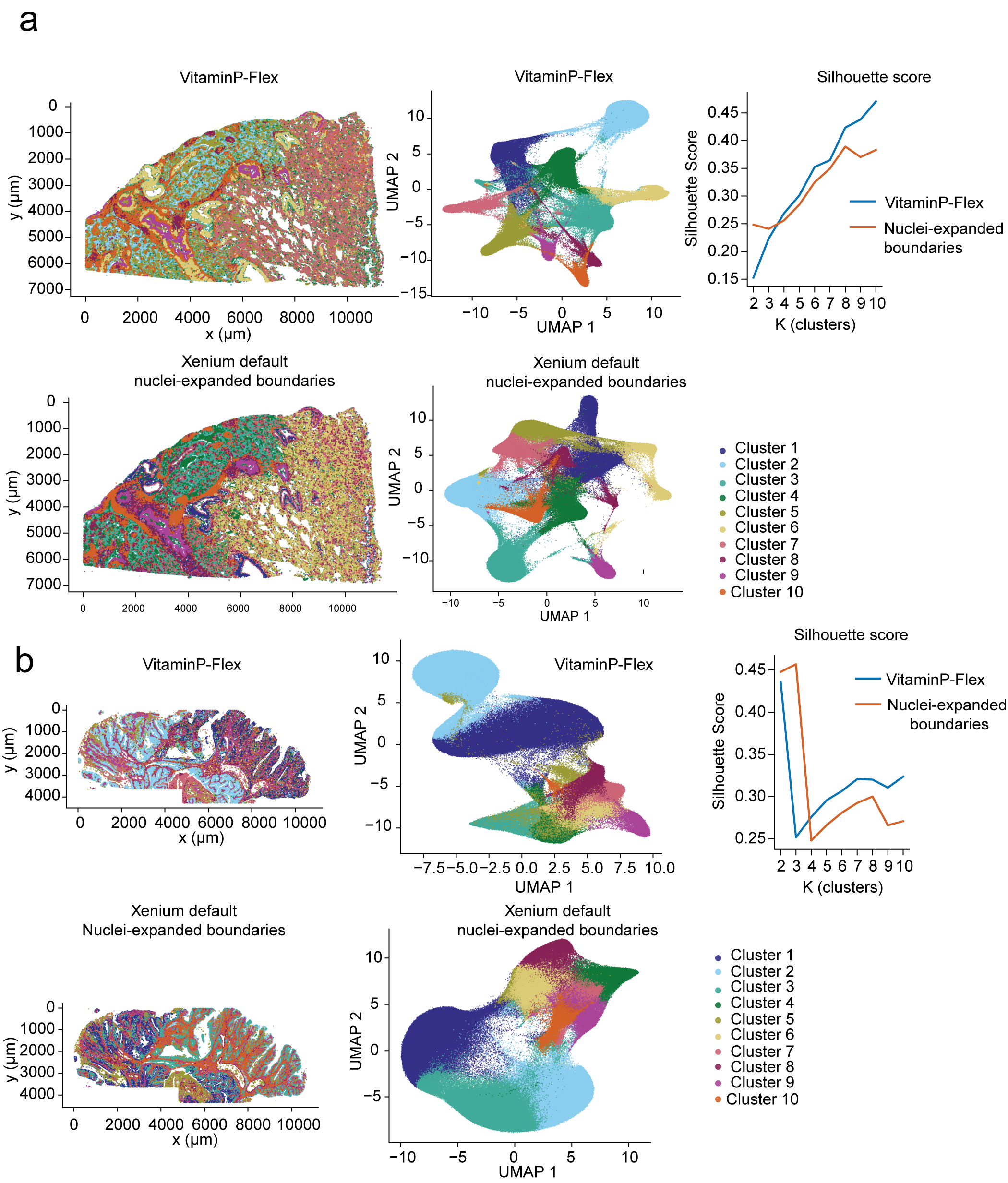}
}

\captionsetup{labelformat=empty}
\captionof{figure}{
\textbf{Extended Data Fig. 4: Comparison of spatial clustering and embedding quality between VitaminP-Flex and Xenium default segmentation across two Xenium datasets.}
\textbf{a,} Lung cancer sample. Spatial clustering and embedding comparison for a lung cancer specimen from the 10x Xenium dataset. Left panels show spatial maps of cells colored by cluster identity derived from VitaminP-Flex segmentation, while the corresponding panels below show clustering obtained using the Xenium default segmentation with nuclei-expanded boundaries. Middle panels show UMAP embeddings of the gene-expression profiles colored by cluster assignment ($K = 10$ clusters). Embeddings derived from VitaminP-Flex segmentation exhibit more compact and better-separated clusters compared with those obtained using Xenium default segmentation, which show increased cluster overlap and mixing. Right panels show clustering quality across resolutions ($K = 2\text{--}10$) evaluated using the silhouette score, where higher values indicate improved cluster separation. Across clustering resolutions, embeddings derived from VitaminP-Flex segmentation consistently achieve higher silhouette scores than those derived from Xenium default segmentation.
\textbf{b,} Colorectal cancer sample. Equivalent comparison performed on a colorectal cancer specimen from the Xenium dataset. Spatial cluster maps (left) show cluster assignments derived from VitaminP-Flex segmentation and Xenium default segmentation, respectively. Middle panels display the corresponding UMAP embeddings, demonstrating improved separation of transcriptionally distinct cell populations when using VitaminP-Flex segmentation. Right panels show silhouette score comparisons across clustering resolutions ($K = 2\text{--}10$), again indicating improved clustering quality for embeddings derived from VitaminP-Flex. Clusters were computed using k-means clustering ($K = 10$) on normalized gene-expression embeddings. A consistent color palette was used to visualize cluster identities across segmentation methods for comparability.
}
\label{fig:ext_fig4}
\captionsetup{labelformat=default}

\noindent\makebox[\textwidth]{%
\includegraphics[width=\textwidth]{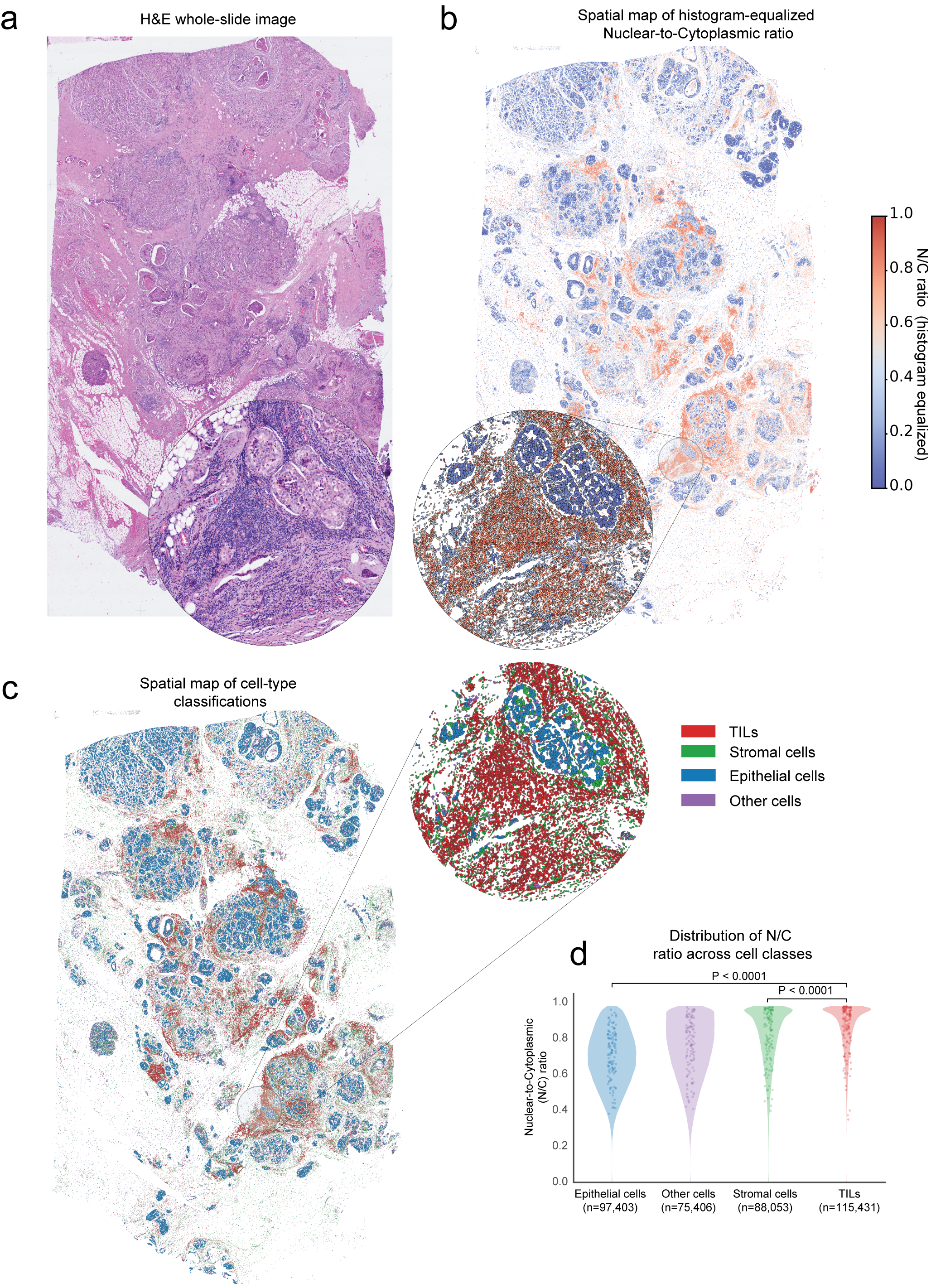}
}

\captionsetup{labelformat=empty}
\captionof{figure}{
\textbf{Extended Data Fig. 5: Whole-slide nuclear-to-cytoplasmic (N/C) ratio mapping and cell-type classification derived from joint nuclear and whole-cell segmentation.}
\textbf{a,} Whole-slide H\&E image of a breast cancer tissue section from the 10x Xenium public dataset. A representative region is highlighted to illustrate local tissue morphology and cellular architecture.
\textbf{b,} Spatial map of per-cell N/C ratios computed across the entire tissue section. Colors represent histogram-equalized N/C values (blue = low, red = high) to enhance visualization of spatial variation. The inset highlights local heterogeneity in N/C ratios across tumor and stromal regions.
\textbf{c,} Spatial map of predicted cell-type classifications across the same tissue section. Individual cells are colored according to predicted class: tumor-infiltrating lymphocytes (TILs), epithelial cells, stromal cells, and other cells. This map illustrates the spatial organization of immune, epithelial, and stromal compartments within the tumor microenvironment.
\textbf{d,} Distribution of N/C ratios across major cell classes. Violin plots summarize the distribution of per-cell N/C ratios for epithelial cells, stromal cells, TILs, and other cells, with embedded boxplots indicating the median and interquartile range. Points represent subsampled individual cells. Consistent with expected morphology, TILs exhibit higher N/C ratios relative to stromal and epithelial cells. N/C ratios were computed from instance-level nuclear and whole-cell segmentations generated by VitaminP. Across the analyzed tissue section, 557{,}469 cells and 492{,}599 nuclei were detected. The median N/C ratio was 0.879 (mean 0.821). Cells without associated nuclei accounted for 12.2\% of instances, while only 344 orphan nuclei ($<0.1\%$) lacked an associated cell instance. The analyzed region covered approximately 48{,}649 $\times$ 73{,}245 pixels, extracted from a 51{,}265 $\times$ 74{,}945 pixel whole-slide image.
}
\label{fig:ext_fig5}
\captionsetup{labelformat=default}

\section*{Patient Mosaic Team}

{\small
\noindent
Nadim J Ajami\\
Azad Ali\\
Franklin Alvarez\\
Brittany Alverez\\
Bianca Amador\\
Surosh Avandsalehi\\
Claudia Alvarez Bedoya\\
Katrice Bogan\\
Elena Bogantenkova\\
Elizabeth Bonojo\\
Maria Neus Bota-Rabassedas\\
Elizabeth M Burton\\
Noble Cadle\\
Vanessa Castro\\
Chi-Wan Chow\\
Randy Aaron Chu\\
Candace Cunningham\\
Carrie Daniel-MacDougall\\
Nana Kouangoua Diane C\\
Mary Domask\\
Sheila Duncan\\
Andrew Futreal\\
Vivian Gabisi\\
Jessica Gallegos\\
Andrea Galvan\\
Ana Garcia\\
Jose Garcia\\
Celia Garcia-Prieto\\
Christopher Gibbons\\
Jonathan Benjamin Gill\\
Dominic Guajardo\\
Curtis Gumbs\\
Kristin J Hargraves\\
Tim Heffernan\\
Joshua Hein\\
Sharia Hernandez\\
Charlotte Hillegass\\
Yasmine M Hoballah\\
Theresa Honey\\
Chacha Horombe\\
Habibul Islam\\
Stacy Jackson\\
Jeena Jacob\\
Akshaya Jadhav\\
Robert Jenq\\
Weiguo Jian\\
Juliet Joy\\
Isha Khanduri\\
Walter Kinyua\\
Laura Klein\\
Mark Knafl\\
Larisa Kostousov\\
Ying-Wei Kuo\\
Wenhua Lang\\
Barrett Craig Lawson\\
Alexander Lazar\\
Jack Lee\\
Erma Levy\\
XiQi 'Cece' Li\\
Latasha D Little\\
Yang Liu\\
Yan Long\\
Vielka Lopez\\
Wei Lu\\
Sandra Lugo\\
Aaliyah Maldonado\\
Jared Malke\\
Asri Margono\\
Dipen Maheshbhai Maru\\
Grace Mathew\\
Brian McKinley\\
Jennifer Leigh McQuade\\
Courtney McRuffin\\
Gertrude Mendoza\\
Christopher Miller\\
Raymond Montoya\\
Francisco Montemayor\\
Theresa Nguyen\\
Heather Perez\\
Juan Posadas Ruiz\\
Sabitha Prabhakaran\\
Mallory Psenda\\
Gabriela Raso\\
Mike Roth\\
Pranoti Sahasrobhojane\\
Amber Savant\\
Keri L Schadler\\
Alejandra Serrano\\
Kenna R Shaw\\
Julie M Simon\\
Elizabeth Sirmans\\
Luisa Maren Solis Soto\\
Xingzhi 'Henry' Song\\
Meghan Stennis\\
Huandong 'Howard' Sun\\
Maria Chang Swartz\\
Marialeska Tariba-Edick\\
Christopher Vellano\\
Angela Walker\\
Ignacio Ivan Wistuba\\
Scott Eric Woodman\\
DeArtura Young\\
Jianhua 'John' Zhang\\
Haifeng Zhu\\
Hui 'Helen' Zhu\\
Olga Bat\\
Shadarra Crosby\\
Ellie Freebern\\
Cindy Hwang\\
Diana Kouangoua\\
Yang Li\\
Sharon Miller\\
Xiaogang 'Sean' Wu\\
}

\end{document}